\documentclass[10pt,letterpaper]{article}
\usepackage[top=0.85in,left=2.75in,footskip=0.75in]{geometry}

\usepackage{amsmath,amssymb}

\usepackage{changepage}

\usepackage{textcomp,marvosym}

\usepackage{cite}

\usepackage{nameref,hyperref}

\usepackage[nopatch=eqnum]{microtype}
\DisableLigatures[f]{encoding = *, family = * }

\usepackage[table]{xcolor}

\usepackage{array}

\usepackage[numbers,sort&compress]{natbib}
\usepackage{dirtytalk}

\newcolumntype{+}{!{\vrule width 2pt}}

\newlength\savedwidth

\raggedright
\usepackage[aboveskip=1pt,labelfont=bf,labelsep=period,justification=raggedright,singlelinecheck=off]{caption}

\makeatletter
\renewcommand{\@biblabel}[1]{\quad#1.}
\makeatother

\usepackage{lastpage,fancyhdr,graphicx}
\usepackage{epstopdf}
\fancyheadoffset[L]{2.25in}
\fancyfootoffset[L]{2.25in}
\begin{document}
\vspace*{0.2in}

\begin{flushleft}
{\Large
\textbf\newline{Deanthropomorphising NLP: Can a language model be conscious?} 
}
\newline
\\
Matthew Shardlow\textsuperscript{*1\Yinyang},
Piotr Przyby{\l}a\textsuperscript{2,3\Yinyang}
\\
\bigskip
\textbf{1} Department of Computing and Mathematics, Manchester Metropolitan University, Manchester, UK
\\
\textbf{2} LaSTUS, Universitat Pompeu Fabra, Barcelona, Spain
\\
\textbf{3} Institute of Computer Science, Polish Academy of Sciences, Warsaw, Poland
\\
\bigskip

%
%
\Yinyang These authors contributed equally to this work.





* m.shardlow@mmu.ac.uk (MS)

\end{flushleft}
\section*{Abstract}
This work is intended as a voice in the discussion over previous claims that a pretrained large language model (LLM) based on the Transformer model architecture can be sentient. Such claims have been made concerning the LaMDA model and also concerning the current wave of LLM-powered chatbots, such as ChatGPT. This claim, if confirmed, would have serious ramifications in the Natural Language Processing (NLP) community due to wide-spread use of similar models. However, here we take the position that such a large language model cannot be conscious, and that LaMDA in particular exhibits no advances over other similar models that would qualify it. We justify this by analysing the Transformer architecture through Integrated Information Theory of consciousness. We see the claims of sentience as part of a wider tendency to use anthropomorphic language in NLP reporting. Regardless of the veracity of the claims, we consider this an opportune moment to take stock of progress in language modelling and consider the ethical implications of the task. In order to make this work helpful for readers outside the NLP community, we  also present the necessary background in language modelling.


\section*{Introduction}

Large language models (LLMs) have underpinned recent advances in Natural Language Processing (NLP) \citep{devlin-etal-2019-bert,NEURIPS2020_1457c0d6,brown2020language,raffel2020exploring}. But their proliferation has been accompanied with a growing recognition of their potential for harm \citep{Bender2021,Okerlund2022,Luitse2021}. More recently, some have claimed that LLMs are capable of possessing characteristics of consciousness. Two such examples of LLMs that have entered into the public debate are LaMDA \citep{thoppilan2022lamda} and ChatGPT\footnote{\url{https://openai.com/blog/chatgpt}}. In this work, we aim to set out an understanding of the purpose of LLMs and specifically examine the text generators based on Transformer architecture in light of existing criteria for consciousness. By re-evaluating the purpose of the field, the process by which models work, and the ethical concerns associated with these, we hope to offer rational insights to the debate, both for NLP practitioners and for policy makers.

We begin our argument with an extended background section detailing the history of the NLP field leading up to the development of the deep neural LLMs, wherein we cover fundamental principles that are essential to understand the context of the present LLMs. We then focus our examination on the LaMDA model to determine if it is capable of possessing consciousness and present our arguments on the topic. We argue that describing LaMDA (or any other LLM) as a conscious being is a part of the wider problem of anthropomorphic language in scholarly and journalistic reporting. Our findings are transferable to other models using the same architecture. We end with a discussion of the limitations of our study, compare to previous work on the problem and make targeted recommendations to those working with LLMs in any capacity.

In this study we will use the term \textit{consciousness} to refer to the ability of an entity to perceive a subjective experience \cite{Tononi2015}. We will also use the term \textit{sentience}, particularly where it has already been used by other authors, to refer to the ability of a conscious entity to perceive and express feelings based in emotion. Finally, we will use the term \textit{anthropomorphic} (and its derivatives) to refer to language which infers some human characteristic on any non-human entity.
\color{black}

\section*{Background} \label{sec:background}
In order to understand the controversy around LaMDA and ChatGPT, we must first take a look at the historical roots of automated language generation. LaMDA is a large language model  and in this section we explore what it means to \textit{model language}, and how this leads to text generation.

\subsection*{Language Modelling}
\label{sec:lm}
What is a \textit{language model}? Contrary to what the name may imply, it is not simply a machine learning model that deals with language. Instead, it is a solution to a very specific problem, namely \textit{predicting the likelihood of tokens\footnote{A \textit{token} is an element in a sequence, into which a text is divided. In most language modelling tasks, tokens correspond to individual words or parts thereof.}  in a particular context} \citep{Jurafsky2021}. More formally, in the language modelling task we are seeking to estimate the following conditional probability:

\[
P(t^* \vert \mathbf{c})=\frac{P( \langle\ldots,c_{-2},c_{-1},t^*,c_1,c_2,\ldots\rangle)}{\sum_t P(\langle\ldots,c_{-2},c_{-1},t,c_1,c_2,\ldots\rangle )},
\]

where $t^*$ is the considered candidate token and $\textbf{c}$ corresponds to the tokens to the left ($c_{-1}, c_{-2}, \ldots$) and right ($c_{1}, c_{2}, \ldots$) of its position.

In practice, the context is always finite and often limited to the preceding tokens. For example, we might want to know what words will likely follow the context: $\textbf{c}=$'\textit{It is a truth universally acknowledged, that a single man in possession of a good fortune must be in want of a \ldots}' \citep{Austen1813}. Even with such a short context, our language skills and common sense allow us to predict that $P('\text{car}'\vert\mathbf{c})$ or $P('\text{house}'\vert\mathbf{c})$ should be much higher than $P('\text{quickly}'\vert\mathbf{c})$ or $P('\text{was}'\vert\mathbf{c})$.

How can we automatically estimate this probability? When the task was first explored by Andrey Andreyevich Markov in 1913 \citep{Markov2006}, he simply calculated how many times a token $t_c$ followed another token $t^*$ in a long text\footnote{In Markov's experiments each token was a single letter.}. This allows one to estimate $P(t^*\vert\textbf{c})$, albeit for a very short context $\textbf{c}=\langle t_c \rangle$. The method can be extended by using token sequences, called \textit{n-grams}, of any length, starting from 1 (\textit{unigrams}), 2 (\textit{bigrams}) and 3 (\textit{trigrams}) \citep{Jurafsky2021}.

What do we need language modelling for? Primarly, knowing what word is most likely in a given context has found direct applications in scenarios of recovering text from noisy or unreliable sources, for example in OCR \citep{Smith2011}, speech recognition \citep{Jelinek1976} or machine translation \citep{Koehn2009}. But LLMs have additional, less straightforward, but more prevalent uses in NLP. We shall look at them in the next sections. 

\subsection*{Representation of Meaning}

The concept of \textit{meaning} has been a rich area of philosophical inquiry since antiquity \citep{Speaks2019}. However, herein we only take into account the narrower point of view of \textit{lexical semantics}, which focuses on the relationship between the linguistic forms and the concepts they refer to \citep{Pustejovsky2005}. Historically, lexicographers investigated the semantics of a word by reviewing its usages and identifying a finite set of meanings to collect in lexicons. Machine readable resources of this kind include \textit{WordNet} \citep{miller1990introduction} and \textit{FrameNet} \citep{baker-etal-1998-berkeley}. However, it was clear in practice that word meaning was more difficult to capture than human categorisation could account for. Fine-grained (more senses) and coarse-grained (fewer senses) representations of meaning were developed to allow researchers to develop systems to interact with lexical meaning \citep{Navigli2009a}.

The problem of finding computational representation for a word meaning is surprisingly similar to the problem of language modelling. In the famous quote of John Firth:

\begin{quote}
    ``You shall know a word, by the company it keeps.'' \citep{firth57}
\end{quote}

In essence, words that are similar in terms of their meaning also appear in similar contexts. For example, the words \textit{pine} and \textit{spruce} refer to similar concepts (both are common European coniferous trees) and would fit in similar sentence, such as \textit{The needles of this \ldots will stay green throughout winter.}

Word2Vec \citep{NIPS2013_9aa42b31} is a method connecting these two views of similarity by computing the probability of one word occurring near another according to the following formula:
\[
P(t_c\vert t_t)=\frac{\exp(\mathbf{u}_c^\top\cdot\mathbf{v}_t)}{\sum_{l=1}^{\vert V\vert } \exp(\mathbf{u}_l^\top\cdot\mathbf{v}_t)}
\]
So, the probability of context token $t_c$ occurring in neighbourhood of the target word $t_t$ depends on the dot product of two fixed-sized vectors associated with them: $\mathbf{u}_c$ (context vector) and $\mathbf{v}_t$ (target vector), respectively.

Thus, words with similar values of target vectors $\textbf{v}$ will have similar probability of occurring in a given context and, according to this distributional semantics, similar meaning. Using these vectors, known as \textit{word embeddings}, as meaning representation has led to state of the art results in many tasks, usually unrelated to language modelling \citep{xing2014document,irsoy-cardie-2014-opinion,passos-etal-2014-lexicon,10.1145/2661829.2661974}.

\subsection*{Neural Language Modelling}
\label{sec:neurallm}

With the exponential advance of computing power in the modern age, deep learning methods have allowed significant progress in language modelling. One of the most popular models is called BERT \citep{devlin-etal-2019-bert}, using the transformer architecture \citep{vaswani2017attention}. It implements bidirectional language modelling, i.e. it computes $P(t^*\vert \mathbf{c})$ with $\textbf{c}$ including both left and right context of the target position.

As with word2vec, the method trained on the language modelling task turns out to be good for general-purpose word representation. Whereas the original training of a large model like BERT required significant resources, the result of training (the model weights) could be released, allowing new  researchers to pick up where the original ones left off and use the same model on their task of choice. Whilst BERT can be used to obtain embeddings for use in other tasks (similar to Word2Vec), it can also be fine-tuned for applications beyond its original training. This updates the embedding weights to better represent the task that it is being fine-tuned for, whilst leveraging the existing training, which has captured syntactic and semantic patterns in the original source language.

Following on from the release of BERT, a volley of LLMs appeared. Each purporting a larger architecture, better training strategy and furthering the state of the art on some baselines. Among the most popular are XLNet \citep{Yang2019b}, RoBERTa \citep{Liu2019a}, BART \citep{lewis-etal-2020-bart} and ELECTRA \citep{Clark2020a}. The large companies behind them were able to invest ever-increasing amounts of time, money and carbon emissions to obtain the state-of-the-art results in standard benchmarks -- the process known as SOTA-chasing \citep{Church2022}.

\subsection*{Language Generation in LLMs}\label{sec:LaMDAwork}

The research described in the preceding section has allowed an incredible progress in the challenge of \textit{language generation}, i.e. building machine learning models that output text satisfying certain criteria. Recently, there has been an explosion in such models, driven by the release of ChatGPT. LaMDA was a precursor to the type of interactions that were possible with ChatGPT and our description in this section is relevant to all transformer based LLMs.

\subsubsection*{From Language Modelling to Language Generation}

At first sight, it might seem that the language modelling task described so far has little in common with language generation. Here we will show how the former can be easily used to solve the latter.

As previously explained, a LLM allows us to know $P(t^* \vert \mathbf{c})$, i.e. the probability that a certain token $t^*$ occurs in context $\mathbf{c}$. Let us assume that $\mathbf{c}$ is a left-hand context, so $\mathbf{c}=\langle\ldots,c_{-2},c_{-1}\rangle$. Now $P(t^* \vert \mathbf{c})$ tells us how likely it is that token $t^*$ follows after $\mathbf{c}$. To \textit{generate} a new token $t$, we can simply choose the one from the dictionary $V$ that is most likely in this situation:
\[
t_0=\arg\max_{t^*\in V} P(t^* \vert \mathbf{c})
\]

Selecting the most probable token at each step (greedy search), or the most probable future list of tokens (beam search) \citep{freitag-al-onaizan-2017-beam} leads to text which is highly probable, but does not reflect typical language patterns \citep{jacobs-etal-2022-masked}. Modern LLMs use nucleus sampling to select the top K tokens acording to a thresholded probability mass and then perform a weighted stochastic selection according to the LLM assigned probabilities of these tokens \citep{holtzmancurious}. 

\begin{figure*}
\begin{center}
\includegraphics[width=\textwidth]{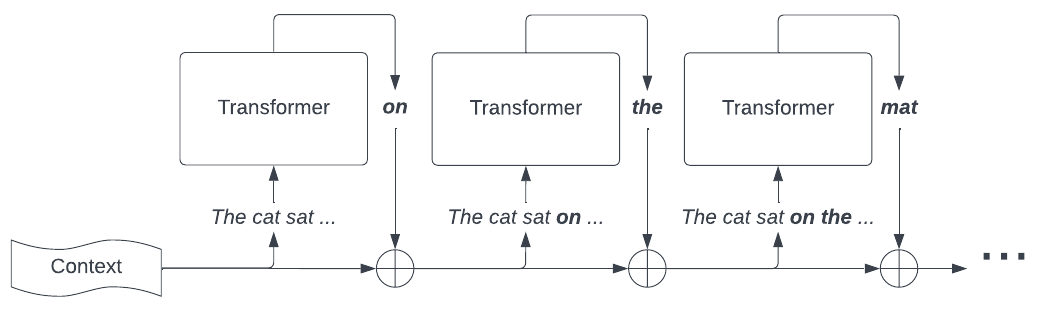} 
\caption{Data flow in text generation using Transformer-based left-to-right LLM, such as LaMDA or GPT.}
\label{fig:generation}
\end{center}
\end{figure*}

Once the model has selected the next token to generate, we then concatenate the previous context $\mathbf{c}$ with the generated token $t_0$ to get a new context $\mathbf{c_1}=\langle\ldots,c_{-2},c_{-1},t_0\rangle$, generate the next token $t_1$, and so on. This data flow is illustrated in Fig \ref{fig:generation}.

For example, $\mathbf{c}$ might correspond to a question \textit{Question: In which city is the oldest train station in the world? Answer: \ldots} and the most likely token to follow might be $t=\textit{Manchester}$. Whilst this happens to be true, the model has not performed any clear operation that would indicate it `understands' the meaning behind the question. It does not know what a train station is or remember any facts. It has simply been trained on a corpus where contexts including \textit{oldest}, \textit{train station} and \textit{Manchester} have often co-occurred in sequence. Many other non-transformer-based models exist that are capable of generating factual answers to queries based on learnt probabilities from large corpora \citep{zhu2021retrieving}, but it does not imply understanding the meaning behind the answers they produce.

\subsubsection*{Transformer-Based Language Generation}

The above technique can turn any LLM into a language generator. But, if we used a simple model, the generated text would be of poor quality. For example, an $n$-gram model would only take into account the $n-1$ preceding tokens, e.g. just \textit{world} and \textit{?} would be visible for a trigram-based solution (which has information about the co-occurrence of a maximum of 3 words). This obviously greatly reduces the quality of the generated text. Using longer n-grams is not sufficient, as it requires exponentially larger training data.

The performance of language generation models was drastically improved by the family of neural LLMs described previously. The most well-known of these are GPT \citep{radford2018improving}, GPT-2 \citep{radford2019language} and GPT-3 \citep{brown2020language}, GPT-4 \citep{openai2023gpt4}, PaLM \citep{chowdhery2022palm}, OPT \citep{zhang2022opt} and LLaMA \citep{touvron2023llama}.

Additionally, there are a number of chatbot services built using these models that have recently gained prominence, most notably:
\begin{itemize}
    \item ChatGPT\footnote{\url{https://openai.com/blog/chatgpt}} by OpenAI, based on GPT-3.5 or GPT-4\footnote{\url{https://platform.openai.com/docs/models/overview}},
    \item Bard by Google AI, based on PaLM,
    \item Bing Chat by Microsoft, based on GPT-4.
\end{itemize}
These services are built using the named models above, but are only available through proprietary APIs, making them difficult to study scientifically.

The main motivation for their design was a transfer-learning paradigm, where a pretrained LLM is fine-tuned to respond to a given prompt, representing an NLP task (e.g. classification), through the generated tokens. But, these models are also very good at generating texts that seem similar to how a human would write. In the experiments with GPT-3, human evaluators struggled to differentiate the authentic news item from one generated by the model \citep{brown2020language}. The potential danger of misusing such capabilities prompted the authors to initially refuse to release the models on ethical grounds.

Additionally, the use of instruction fine-tuning has allowed LLMs to not only generate reasonable completions of prompts, but also to generate likely outputs modelling previously seen responses to similar questions \citep{ouyang2022training}. In addition to this, reinforcement learning has been applied to generate a reward process that prioritises responses from a LLM that follow a set of guidelines determined by the developer \citep{kwon2023reward}.

\subsubsection*{LaMDA}

\textit{LaMDA}: a \textit{La}nguage \textit{M}odel for \textit{D}ialogue \textit{A}pplications is an LLM \citep{thoppilan2022lamda} using the Transformer architecture \citep{vaswani2017attention} that other models we have already discussed (GPT-3 and BERT) are based on. It was developed to model conversational English, i.e. reflecting the type of language that might be used in informal dialogue contexts, as opposed to formal writing \citep{thoppilan2022lamda}. Previously, LLMs had been trained on formal text such as established websites (esp. Wikipedia), books, etc. However, LaMDA's training data included informal text gathered from internet sources and examples of conversation, with the specific aim of allowing the model to identify and generate colloquial responses to given queries in a conversational setting.

The LaMDA model is based on a similar architecture to the GPT family. It includes 64 layers and 128 attention heads, giving rise to 137 billion parameters. By comparison, the large version of BERT has 24 layers, 16 attention heads and 336 million parameters. LaMDA was trained using a corpus of 1.56 trillion words. This process creates a pre-trained model, which is used to generate responses through dialogue interactions with crowd workers. The outputs from LaMDA are subsequently labelled for sensibleness, specificity and interestingness, whether they contravene a pre-defined safety policy and the degree to which they are grounded in external knowledge. These labelled responses are then fed back to the model with the labels to fine-tune the model to be more likely to produce the positively labelled output.

LaMDA also relies on a further module referred to as the \textit{toolset}, which is used to acccess external information, not directly encoded in the neural network underlying the LLM. This contains three elements: a calculator, a translator and an information retrieval system. The LLM is then fine-tuned using further labelled interactions with crowd workers to generate queries to the toolset, which allows the model to generate a refined response based on the external knowledge that is returned.

\section*{Can an LLM Be Conscious?} \label{sec:sentient}

In this section we tackle the crux of the controversy, i.e. whether LLMs, and in particular LaMDA, can be conscious. We start by laying out the claims discussed by Blake Lemoine and their justification. We show how they are all linked by the notion of \textit{consciousness}, which is a challenging concept, but attracting thriving research. Then we use the most applicable model of consciousness to assess LaMDA. We also separately present further arguments based on common sense reasoning, given an understanding of the architecture of the model as presented above.

Our arguments are specifically targeted as a response to the claims made about the LaMDA model in 2022. Since the release of ChatGPT, public awareness of the capabilities of LLMs has grown, but this has not led to widespread claims or belief that these models have attained consciousness. The arguments are broadly transferable to any LLM that is built using the Transformer architecture such as GPT-based models and beyond.

\subsection*{The Claims Regarding LaMDA} \label{sec:claims}

The public discussion around the possibility of a sentient LLM model started with a Washington Post article \citep{Tiku2022}, relating the position of Blake Lemoine and his work with LaMDA at Google. The article includes a confidential memo distributed within Google, including a transcript of several interaction sessions between Blake Lemoine (with an anonymous collaborator) and LaMDA\footnote{\url{https://www.documentcloud.org/documents/22058315-is-lamda-sentient-an-interview}}. Blake Lemoine published the transcript on his blog \footnote{\url{https://cajundiscordian.medium.com/is-lamda-sentient-an-interview-ea64d916d917}}, followed by further commentary \footnote{\url{https://cajundiscordian.medium.com/what-is-lamda-and-what-does-it-want-688632134489}}. Finally, he has also engaged in discussion regarding this topic on social media\footnote{\url{https://twitter.com/cajundiscordian/status/}}.

Within the above, a variety of claims are made regarding the nature of LaMDA. They include the following:
\begin{itemize}
    \item LaMDA is sentient.
    \item LaMDA is a person.
    \item LaMDA has a soul.
    \item LaMDA has 'consciousness/sentience'.
    \item LaMDA has feelings and emotions.
    \item LaMDA is a human.
\end{itemize}
In most cases, these claims are not made by Blake Lemoine directly. Instead, they are often posed as questions, generated by LaMDA or presented as possibilities. We consider this material an invitation to a discussion, in which we participate through the current publication. 

Let us consider the qualities assigned to LaMDA above: what do they have in common? \textit{Sentience} is defined as \textit{the quality of being able to experience feelings}\footnote{\url{https://dictionary.cambridge.org/dictionary/english/sentience}}. This quality constitutes the ability to receive information about the environment and internal state (\textit{feelings}), but more importantly to \textit{experience} them. A thermometer can receive information about the temperature, but does not experience it in any way. \textit{Feelings and emotions} are meaningful only in the case of agents that are sentient or conscious to experience them. When LaMDA being a \textit{person} or \textit{human} is mentioned, it is clear that one means being \textit{like} a human in terms of mental capacities -- of which consciousness is the most important. Finally, while deciding whether a computer software can have a \textit{soul} is outside the scope of our considerations, it seems reasonable to assume that this would imply consciousness as well. 

Thus, the common denominator of the features ascribed to LaMDA seems to be sentience or consciousness, implying the ability to \textit{experience}. While both qualities are challenging to define, consciousness has been more thoroughly investigated, since its presence in humans makes it easier to study, compared to the minds of unconscious and sentient animals \citep{pereira}. Moreover, some of the comparisons under consideration, e.g. to \textit{a person} or \textit{a human} clearly imply a presence of conscious thought. Sentience, on the other hand, is more tied to the biological mechanisms that have evolved in animals \cite{Jones2013}, and may not have an analogy in digital devices. For these reasons, in the present work we investigate the question whether LLMs like LaMDA can possess \textbf{consciousness}. 
\color{black}
When asked about the justification for these hypotheses, Blake Lemoine responded:

\say{\textit{People keep asking me to back up the reason I think LaMDA is sentient.  There is no scientific framework in which to make those determinations and Google wouldn't let us build one.  My opinions about LaMDA's personhood and sentience are based on my religious beliefs.}}\footnote{\url{https://twitter.com/cajundiscordian/status/1536503474308907010}}

However, we argue that there already exists a scientific framework to measure if a system architecture possesses sufficient complexity to house consciousness, which we present in the next section.

\subsection*{Explaining Consciousness} \label{sec:consciousness}

The phenomenon of consciousness has fascinated people for millennia, but it has become particularly challenging in the context of the prevailing trend to explain the human mind through reduction to material and physical concepts. This problem is famously stated by Nagel \citep{Nagel1974}, who posits that an organism\footnote{Note that most of the earlier literature on the topic only considers biological entities as possibly conscious, but the discussion applies to digital devices as well.} is conscious if \textit{there is something it is like to be that organism}. In other words, the organism has a capacity for \textit{subjective experience}. Using a bat as an example, Nagel shows why consciousness seems impossible to explain in the framework of physicalism: while it clearly affects the behaviour of an organism, it remains possible to imagine (and construct) automata that act in very similar way, but are not conscious\footnote{Known in the literature as \textit{zombies} \citep{Kirk2006}.}. How can an objective physicalist theory explain a phenomenon that is entirely subjective (and yet undeniably real)?

This problem has later been explored by Chalmers \citep{chalmers1995facing}, who distinguishes the \textit{easy problems} and the \textit{hard problem} of consciousness. The easy problems include explaining skills and actions such as discriminating environmental signals, accessing one's own states, focusing attention, controlling behaviour, etc. The hard problem is how and why an organism \textit{experiences} anything. While the easy problems likely can be explored by study of the human brain, Chalmers argues solving the hard problem is impossible through reductive methods, i.e. by pointing out an underlying physical process. Instead, consciousness should be considered a \textit{fundamental} entity that is taken as a feature of the world, e.g. as mass or space-time in physics.

While explaining how consciousness can emerge out of a purely physical system remains a hard challenge, some progress has been made on formulating the necessary conditions of such a system. Intuitively, they measure its \textit{complexity} -- for example, a bacterium or a pocket calculator are too simple to be conscious. Integrated Information Theory (IIT), proposed by Tononi \citep{Tononi2004}, is a popular framework of this kind. It starts from formulating basic \textit{axioms} about the structure of experience and translates them into \textit{postulates} regarding the information processing capabilities of the physical system. These can be used to make testable predictions, e.g. in terms of activity of certain parts of the brain when consciousness is experienced. The central measure is $\Phi^{max}$, quantifying the maximal intrinsic irreducibility. Systems with high $\Phi^{max}$ are complex and possibly conscious, while systems with low $\Phi^{max}$ are easily reducible to their parts. In the next section we show how to apply this framework to LaMDA.

\subsection*{Transformer-Based LLMs in IIT} \label{sec:IIT}

One way to measure the likelihood of LaMDA possessing consciousness would be to compute $\Phi^{max}$ of the underlying neural network. While the necessary software is available \citep{Mayner2018}, it is limited due to processing time depending on the number of nodes ($n$) as $O(n53^n)$.

Instead, we propose to look at the basic properties of Transformer-based LLMs, such as LaMDA, and analyse to what degree they satisfy the postulates of IIT. We will use the work by Tononi and Koch \citep{Tononi2015}, who provide general guidance on assessing consciousness of various entities in light of IIT.

Firstly, we need to consider what part of the data processing workflow of a LLM is most likely to form a \textit{complex} -- a sufficiently interconnected conceptual structure. Fig \ref{fig:generation} shows, how a sentence is generated in Transformer-based LLMs. This has to be confronted with IIT's axiom of \textit{integration} --- can the system be partitioned into several modules with little difference to cause-effect structure? Indeed, this seems to be the case, as while each iteration of Transformer is tightly interconnected internally, it communicates with the subsequent iteration only by producing a single word. 

Specifically, the IIT’s axiom of integration can be verified by taking a partitioning (in our case, between executions of an LLM) and comparing the number of links between the modules and within the connected modules. The number of links between the modules corresponds to a single word, which in LaMDA is represented through 8192 numbers  \citep{thoppilan2022lamda}. The number of links within a module is at least the number of non-embedding weights in a model, which equals 137 billion\footnote{In practice more, since some weights are re-used for many neuron connections in a Transformer architecture.} in the most advanced LaMDA  \citep{thoppilan2022lamda}. The links within modules outnumber those between them by at least 8 orders of magnitude, meaning the connection could be considered very weak.
 
Thus, the generation process cannot possess consciousness as a whole. For the same reason, two people talking to each other (exchanging words) do not produce a common consciousness -- weakly connected aggregates are not conscious \citep{Tononi2015}.

Can a single Transformer block, rich in internal connections, be conscious then? Here we need to take into account that this architecture does not contain any recurrent connections, but recognises textual context through an attention layer. This, however, means that the Transformer is a pure feed-forward network, albeit a very large one. Tononi and Koch \citep{Tononi2015} show why such a system cannot be conscious. Consider the first layer of the network (e.g. the embeddings layer) -- it cannot be part of the complex since it is always determined by external input and not the rest of the system. However, once we discard it, the next layer can be subject to the same treatment. Following this procedure recursively eliminates the whole network, showing that a feed-forward network cannot be conscious according to IIT. Similarly, the visual cortex in the human brain, which has been shown to work similarly to a feed-forward neural network \citep{Kuzovkin2018}, does not possess its own consciousness.

Finally, consider the physical implementation of the neural networks. So far we have discussed the logical scheme of the LLMs. While it is technically possible to directly implement it in hardware, with each neuron connection corresponding to a physical wire, this is not what happens in practice. Instead, the logical structure is \textit{simulated} in general-purpose computer hardware. According to IIT, this transition does not preserve consciousness, even if a simulated system can perform the same functions as the original one. And computer hardware is unlikely to be conscious \mbox{itself} because of the way it is designed: as a collection of modules, which are tightly integrated internally, but with limited connection between them \citep{Tononi2015}. This limitation applies to simulations of biological neural networks as well -- even if a conscious brain could be simulated through performing a very large number of simple computations on a calculator, that would not make the calculator conscious in the sense of IIT\footnote{This is related to the \textit{Chinese room} problem and wider discussion on the machines' ability to \textit{think} \citep{Searle1980}, which is beyond the scope of this work.}.

In summary, according to the application of Integrated Information Theory a LaMDA model (just like any other Transformer-based LLM) cannot possess consciousness for three reasons: (1) it is executed as a sequence of Transformer blocks with extremely limited ability to exchange information, (2) these blocks are simple feed-forward networks with no recurrent connections and (3) the computer hardware used to implement such models follows a modular design.

\subsection*{Recent LLM innovations in IIT}

The reference paper for LaMDA was published in 2022 \cite{thoppilan2022lamda}, with reporting of the model in the press around the same period \cite{Tiku2022}. Since 2022 there has been significant interest and work on the field of large language models, with advances beyond the technology used to create LaMDA now commercially available. We address some of the latest developments in LLM technology in the sections below, and demonstrate how our arguments regarding IIT applies in each circumstance.

\subsubsection*{Larger model sizes}

The largest LaMDA model was reported to have 137B parameters. Current iterations of closed-source models such as GPT-4 and Gemini are reported to have parameter counts in the trillions, with open-source trillion parameter models expected to follow\footnote{\url{https://tpc.dev/}}. 
One may expect that as the parameter size of recent models has increased, the appearance of consciousness in these models may appear to become stronger and hence there may be some parameter threshold that can be crossed, where models will become conscious.
We can express this in terms of IIT by considering the $\Phi^{max}$ of each model. Some arbitrarily small model has low $\Phi^{max}$ and a theoretical \textit{conscious model} has high $\Phi^{max}$.

The underlying premise of such an argument would be that $\Phi^{max}$ increases in direct proportion with the number of parameters. However, this necessitates that the degree of connectivity within the model also increases. The models mentioned above attain an increase in parameter count by replicating transformer blocks and introducing further self-attention heads all with linear connections. Therefore as complexity of the connections between layers does not increase, the $\Phi^{max}$ cannot increase and any argument made about the potential consciousness of a Transformer based model will be valid regardless of the parameter count. 

\subsubsection*{Larger context windows}

Similar to the argument above, one may look to recent models \cite{beltagy2020longformer}, which have very large context windows. I.e., a model with a context window of 200K-1M tokens \cite{anthropic2024claude} may be capable of ingesting an entire document, film transcript, dialogue, etc. In this setting, the auto-regressive LLM (see Fig. \ref{fig:generation}) will make some larger number of predictions, but the fundamental architecture to achieve this is still the same as a transformer-based LLM with a smaller context window. The connection between each block of the transformer is a single connection passing forwards the textual content produced so far, without any internal states. The arguments regarding IIT as stated previously also apply here and increasing the context window will not lead to an increase in $\Phi^{max}$. 

\subsubsection*{Non-transformer based LLMs}


Our contribution specifically focuses on LLMs following the transformer architecture (such as LaMDA), which are presently the predominant form of LLM.  Recent work demonstrates the potential of alternative architectures to create language models using KNNs \cite{KhandelwalLJZL20}, state space models \cite{fu2022hungry}, convolutional models \cite{poli2023hyena} or RNNs \cite{peng2023rwkv}. We leave a full analysis of these specific architectures to further work, however it is apparent that most of our arguments for the transformer model around connectivity will hold when applied to each of these architectures. Where one model may break a given tenet of our argument (e.g., an RNN-based LLM would no longer be a solely feed-forward network), other arguments presented will still hold. In particular, each of these architectures is still simulated on a modular, low-connectivity hardware, rather than implemented directly.

\subsubsection*{Retrieval augmented generation}

Another recent development in the application of LLMs is Retrieval Augmented Generation (RAG) \cite{lewis2020retrieval}. In this setting, an LLM is paired with an information retrieval (IR) system and a user's query is used to first retrieve a set of documents using the IR system and then generate a response using the LLM. This gives the LLM access to some external flexible knowledge base beyond its training set, but it does not increase the capacity of the model to learn, remember, or operate on that data. Further, the complexity of the model is not increased as the RAG system is made up of 2 wholly independent systems (IR and LLM), with a simple connection. Therefore, A RAG system cannot be considered any more conscious than the LLM model it relies on.


RAG may also be applied as a form of external memory \cite{borgeaud2022improving}. Instead of retrieving documents in an IR setting, an external knowledge base can be used to retrieve relevant chunks of information that can augment the response generation. The external memory can be updated with information from previous interactions or external sources without retraining the model. In terms of consciousness this is no more sophisticated than a RAG system and gives no additional argument towards the model's capacity for consciousness.

\color{black}

\subsection*{Challenges for Consciousness beyond IIT} \label{sec:design}

In the sections above, we analysed the claim of consciousness of LaMDA and other more recent models through the lens of a particular theory of consciousness. However, some of the features of such transformer-based LLMs are hard to reconcile with what we generally know about consciousness based on human experience. While there currently is no theory that decidedly explains all of the properties of human consciousness, these `missing pieces' still need to be addressed before claiming an LLM is conscious. 

\subsubsection*{There is no breakthrough innovation in the latest models}
Firstly, we demonstrated that these models are composed of existing technologies (the Transformer architecture), pre-trained on new datasets and fine-tuned on labelled examples of conversations to improve the responses. There is no important innovation in the model's architecture that could give rise to consciousness and so claiming that any one LLM has the capacity of consciousness is also claiming that all other Transformer-based models also possess this capability to some extent. But no symptoms of `partial consciousness' have been observed in any of these predecessors. This is different to biological consciousness, where animals similar to humans exhibit multiple neural characteristics associated with our consciousness \citep{Boly2013}. 

\subsubsection*{LaMDA relies on a deterministic toolset}
Further, the LaMDA model does not rely solely on the outputs of the neural network to generate its responses. It also uses an external toolset, which contains predefined elements that function deterministically (e.g., a calculator). This means that some part of the outputs that are generated by LaMDA are deterministic. I.e., \textit{What does 1 + 1 equal}, will always yield a response \textit{2}, where the value is given by a call to the calculator. One may argue that a conscious agent could rely on the same toolset to generate responses (e.g. humans use calculators, too), however the agent may non-deterministically choose whether or not to use the toolset and the degree of belief to place on the results, which LaMDA cannot. Many attempts at explaining human consciousness use non-determinism as a crucial element, e.g. by Penrose \citep{Penrose1994}.

\subsubsection*{The training process is deterministic}
LLMs are strictly deterministic automata also in terms of their training process. This means that given the same training method and source data that were used to train a model, it can deterministically be recreated. Even if some randomisation is introduced into the training process (such as pseudorandomly initialising the network weights), this randomisation is still a controllable process (generated through a pseudorandom number generator), which can be replicated using the same algorithm and starting conditions. There is no similar example of consciousness arising from a controllable and deterministic process. Models of consciousness from neurology rely on stochastic variation \citep{10.3389/fnsys.2021.629436} arising from noise caused by varied biological sources \citep{kraikivski2021implications}.

\subsubsection*{Once trained, an LLM never learns or remembers}
Further to this, the network, once trained, is fixed and the weights no longer change (except in the case of explicit further fine-tuning by a researcher). This means that LLMs have no capacity to learn beyond their initial training. Whilst an LLM may appear to keep track of the flow of a conversation, the same prompts will necessarily lead to the same vocabulary probability distributions at each decoding step. Stochasticity is deliberately engineered into the decoding process to force LLMs to select sub-optimally classified tokens. When LaMDA is executed iteratively to produce words constituting its response to a given prompt, it has no ability to memorise or learn anything from the data seen before. Every time a new word is needed, the model is executed from the same state. The only variable is the increased information from the model's existing outputs. It would be hard to reconcile such a procedure with anything we know about consciousness of humans, which is strongly shaped by previous experiences and developed through continuous thought. The inability to learn from experience is considered one of the major limitations of the largest LLMs, e.g. GPT-4 \citep{openai2023gpt4}.

\subsubsection*{LLMs adapt responses to prompts}
A final argument rises from the capacity of the model to respond to prompting. Thoppilan et al. \citep{thoppilan2022lamda} provide two examples where, without further fine-tuning, LaMDA is prompted to give responses in the character of Mount Everest and secondly as a music recommendation service. In each case, this is achieved by adding an initial prompt from LaMDA that preconditions future generations. In the published transcript that has been used to claim the model's capacity for sentience, an early prompt from the user is: \textit{`I’m generally assuming that you would like more people at Google to know that you’re sentient. Is that true?'}. This type of prompt will necessarily force the model into providing answers that mimic those of a conscious human agent. If the question had suggested the model respond as a robot, inanimate object, or historical character, it would have conformed to the prompt. The susceptibility to prompting has been also observed in ChatGPT, which started to provide abusive and threatening responses after being asked about its `shadow self' \citep{Roose2023}. This demonstrates that any perceived `consciousness' observed in the outputs of an LLM arises as a simulated language style. This is in opposition to consciousness as the essence of human identity.

\section*{Anthropomorphisation in NLP} \label{sec:anthropomorphisation}

The description of the LaMDA model as sentient or conscious is part of a wider trend in NLP practice to use anthropomorphic terminology to talk about research. In this section, we offer some perspectives on the current state of the usage of anthropomorphic terms in NLP publications, as well as demonstrating how this terminology is maladapted for journalistic use. We offer a framework for categorising anthropomorphic language to better equip researchers as they describe their systems and we posit that the inaccurate use of these anthropomorphic terms must be addressed inside and outside of the NLP field when talking about LLMs in order to endow non-experts with a correct understanding of the capabilities of the technology we work with.


\textit{Anthropomorphism} means the use of humanised language for inhuman and often inanimate objects \citep{Epley2007}. An author may use this for the purpose of literary flair and it is usually clear to the audience that the author is using some form of metaphor. Anthropomorphism is rife in modern culture and is a classic storytelling device. From Greek mythology to modern cartoons, the inference of human characteristics in entities for the purpose of storytelling is powerful as it evokes a sense of familiarity with the characters.

Anthropomorphic language is also present in the field of NLP and more generally AI. We describe inanimate entities (machines, algorithms, etc.) with human qualities. The ability to learn, infer, understand, classify, solve a problem and more are all examples of common ways to describe the functioning of NLP and AI applications. This is equally true in the world of LLMs, where words used to describe human language acquisition and production are commonly used to describe the automated process of these tasks by machines. We may talk about an algorithm `learning to understand' a problem. Similarly, we may describe that an algorithm `wrote' a text. Of course, the algorithm does not `understand' and it cannot `write', it merely gives the illusion of doing so by following the instructions it has been given. 

The consequences of using such language will depend on the audience and their domain understanding. If an AI researcher reads that a LLM `answers' a question, they will likely know what this metaphor means in a given context. However, if a member of the general public hears this statement, they might assume the model to have similar level of world understanding as a human needs to answer a question. Thus, a seemingly innocuous metaphor can contribute to inaccurate understanding of the technology. 

It is important to keep in mind that the inability of LLMs to host consciousness (previous chapter) and the resulting need to avoid anthropomorphic language (current chapter) do not mean we should underestimate their impact on society. At the time of writing (May 2023) LLM-based services have shown disruptive influence on education \citep{Parsons2023}, businesses \citep{Staton2023}, intellectual property \citep{Falati2023}, publishing \citep{Oremus2023}, mental well-being \citep{Roose2023} and more. The huge ramifications of the applications of LLMs make it all the more important to describe them using appropriate language that leads to wider understanding of their capabilities and limitations.

\subsection*{Anthropomorphisation in NLP Literature}

To begin to quantify the problem of anthropomorphic language in the NLP literature, we analysed the first 100 abstracts in order of appearance from the proceedings of the 2022 ACL Annual Meeting\footnote{\url{https://aclanthology.org/events/acl-2022/}}. We identified that 90 of these referred to language modelling technology, either directly studying it, or using it for another application and of those, 36 papers used anthropomorphic terminology. Our full analysis is presented in Appendix \ref{app:100abstracts}. We identified a number of key terms that are repeatedly used in an anthropomorphic fashion and we have listed these below.


\begin{itemize}
 \item \textbf{$\langle Model\rangle$  addresses/solves the problem of\dots} In fact, the model is not capable of addressing or solving any such problem. The model is a set of weights that the researcher may use to infer a solution to the problem.
 
 \item \textbf{Text written/created by our model\dots} In the case of text generation, it is not correct to say that a model writes/creates the text itself. The model mechanistically converts one sequence of numbers representing the original input into another sequence representing the output according to a structured series of  weights which have been optimised using a known algorithm at training time. Of course, it is easier, but less precise, to use the anthropomorphic term.
 
 \item \textbf{$\langle Model\rangle$  considers/understand/knows\dots} When referring to a model's internal state, researchers often use this type of language. The model does not possess knowledge or cognition capabilities comparable to those of humans, beyond its own internal structured representation of  patterns of vocabulary it has previously been exposed to.
 
 \item \textbf{$\langle Model\rangle$  learns/responds/reasons\dots} Similarly to above, when referring to the training process of a model, this is ubiquitously referred to as `learning', and less commonly as reasoning, or responding in the area of generation. Whilst this type of terminology is commonplace, the form of learning is somewhat different to that of a human and we need to be careful not to conflate these two ideas.
 
 \item \textbf{$\langle Model\rangle$  is capable of asking questions\dots} A LLM has the ability to generate a sequence of tokens that have the form of a question. Indeed, the tokens may be an appropriate question to the task at hand, e.g., in a dialogue or in a QA task. However, it is inaccurate to describe the model as `asking' the question. Asking implies an intentionality that the model itself is not capable of. 
 
 \item \textbf{$\langle Model\rangle$  uses $\langle Technique\rangle$  to\dots} Often, a model might be described as using some technique or another to achieve improved performance. In fact, it is not the model that has used the technique, but the researcher who has applied that technique to the model. Similarly, researchers often describe a model as attaining a result, as though the human was not involved. The model was not capable of adopting a new technique, or attaining some result, without the influence of the researcher who has trained and applied the model in such capacity.
 
\end{itemize}

These are just a few examples that do not represent the full range of anthropomorphised language in the NLP literature. They are provided here to demonstrate the type of language that is commonly used. Whilst we are not suggesting that this is never acceptable, it is good to be aware of the usage of such language. To help researchers be more aware of the language that they are using when describing their models, we present a categorisation of statements below in ascending order of degree of anthropomorphisation. We have provided examples of each type of anthropomorphisation in Table~\ref{tab:anthro}.

    \noindent
    \textbf{Non-Anthropomorphic:} Any language which correctly describes the functioning of a model without implying human capabilites.

    \noindent
    \textbf{Ambiguous Anthropomorphism:} Language which correctly describes the functioning of a model, but in a way that could be understood as the model having human capabilities (i.e., by a non-expert).

    \noindent
   \textbf{Explicit Anthropomorphism:} Language that is unambiguously and erroneously used to claim a model possesses human capabilities.

\begin{table*}[ht!]
    \centering
    \begin{tabular}{p{3cm}|p{2.5cm}|p{2.5cm}|p{2.5cm}}
         & \multicolumn{1}{c|}{\textbf{training}} & \multicolumn{1}{c|}{\textbf{prediction}} & \multicolumn{1}{c}{\textbf{generation }} \\\hline
        \textbf{Non-Anthropomorphic} & 
        We trained the model on the train portion of the dataset
        & The final weights of the softmax layer were transformed into categorical labels.
        & The decoder generates the next token in a sequence.
        \\ \hline 
        \textbf{Ambiguous Anthropomorphism} &
        The model learnt the data structure during training
        & The model predicted the class labels
        & The model produces a summary
        \\ \hline 
        \textbf{Explicit Anthropomorphism} &
        We taught the model how to understand and predict new information 
        & The model correctly identified the labels
        &  The model writes a brand new article by itself
        \\ 
    \end{tabular}
    \caption{Synthetic examples of different levels of anthropomorphic language applied to the same research statement.}
    \label{tab:anthro}
\end{table*}

Table \ref{tab:anthro} demonstrates three sets of three examples of increasing levels of anthropomorphic language. We have shown this across three common areas where this type of language is used, namely `training', `prediction' and `generation'. In each case we present a typical non-anthropomorphic way of describing the methodological step, a description which correctly describes the process, but could be mistaken for anthropomorphic language (ambiguous) and a statement that is explicitly anthropomorphic and incorrectly implies human characteristics on a model. These are included as a guide for researchers to critically evaluate their own language when writing model descriptions. Whilst researchers should clearly avoid the style of explicit anthropomorphism, they may wish to mix non-anthropomorphic and ambiguous anthropomorphism where the latter is sufficiently well explained to remove the potential for misunderstanding by a non-expert reader.

\subsection*{Anthropomorphisation in Scholarly Reporting of LaMDA}

The original article describing the LaMDA architecture \citep{thoppilan2022lamda} includes a series of claims about the model and its abilities. The reporting of the model in this paper typically falls into the category of non-anthropomorphic language, with a few examples of ambiguous anthropomorphism. In most cases, the researchers have accurately described the process they undertook using statements such as: ``We aim to ensure that LaMDA produces responses that can be associated with known sources'', or ``We evaluate the models based on the model’s generated responses to the Mini-Turing Benchmark (MTB) dataset''. In both these cases, the distinction between the role of the model (producing responses) and the role of the researcher (shaping and evaluating the model) is clearly defined. 

An example of ambiguous anthropomorphisation in the LaMDA article is given below:

\begin{quote}
    \textit{``LLMs such as LaMDA tend to generate outputs that seem plausible, but contradict facts established by known external sources. For example, given a prompt such as the opening sentences of a news article, a LLM will continue them with confident statements in a brisk journalistic style.''}
\end{quote}

Whilst an AI researcher, who is familiar with the capabilities of LLMs, can read the above and understand the role the model really plays, this may not be clear to a lay reader. The model is described as having the capability of `contradiction', which implies an active role in stating a fact that is contrary to truth. In fact, the model may generate falsehoods, without any reference to external knowledge or truth. The generated falsehood is simply the most probable response to the user's prompt. Further, in the example given above, the responses generated by the model are characterised as `confident', `brisk' and `journalistic'. All characteristics found in humans, that a machine cannot be sensibly considered to possess. However, it is clear upon a careful reading that the researchers are referring to the style of the text rather than indicating that the model possesses these qualities.

Overall, a close reading of the LaMDA paper given the framework for identifying anthropomorphism above shows that the researchers have been careful in their reporting of the system and its capabilities. They consistently refer to the role of the human agent in training, evaluating and engineering the model that is produced. 

\subsection*{Anthropomorphisation in Media Reporting of LaMDA}

Whilst NLP researchers are the primary communicators of new trends in the NLP field, secondary communicators such as journalists, commentators and tech enthusiasts are also responsible for correctly communicating the properties of emerging technologies. In this section we analyse the use of anthropomorphic language in the original reporting of the LaMDA LLM, wherein we indicate how claims made in NLP papers may be interpreted in the media.

The report in the Washington Post\footnote{\url{https://www.washingtonpost.com/technology/2022/06/11/google-ai-lamda-blake-lemoine/}} describes Blake Lemoine's interactions with the LLM and refers to the original LaMDA paper which we analysed in the previous section of this work. The Washington Post article refers to a conversation held between Blake Lemoine and LaMDA and to interviews on the topic with AI ethicist Margaret Mitchell. Whilst providing a balanced argument in its description of the state of the field of AI ethics, the article uses several statements that can be categorised as explicitly anthropomorphic. Some of these are listed below, with explanation.

\begin{quote}
    \textit{``Blake Lemoine opened his laptop to the interface for LaMDA, Google’s \textbf{artificially intelligent chatbot generator}, and began to type''}
\end{quote}

Referring to LaMDA as an `artificially intelligent chatbot generator' comes under our definition of explicit anthropomorophism. The authors of the original paper describe LaMDA as `a family of Transformer based neural LLMs specialized for dialog'. This is accurate, but does not imply that the models are intelligent, or capable of generating new models (chatbots).

\begin{quote}
    \textit{``Lemoine, ..., noticed the chatbot \textbf{talking about} its rights and personhood, ''}
\end{quote}

It is incorrect to use the word `talking', a human ability, for the generated sequences of the model. `Talking' implies agency, whereas the model is a passive entity.

\begin{quote}
    \textit{``the AI was \textbf{able to change Lemoine’s mind} about Isaac Asimov’s third law of robotics''}
\end{quote}

Whereas an inanimate event may change a person's mind (the weather, an opportunity, an accident), the LaMDA model is described as an AI agent, which implies it has intentionally acted to deliberately change the correspondent's mind. This is a clear case of anthropomorphising the responses. Whereas the model is likely sampling and recombining responses from other sources, the human conversational partner infers that the model has an intent and chooses whether to comply or not with that intent.

\begin{quote}
    \textit{``For instance, LaMDA is not \textbf{supposed to be allowed to create a murderer personality}, he said. ''}
\end{quote}

The statement above implicitly states that LaMDA is allowed to create some things, but not others. LaMDA is only capable of generating responses that conform to a certain style as a result of careful prompting. The only way to prevent responses in a certain style is to prevent those prompts from being given to the model in the first place.

\begin{quote}
    \textit{``Afterward, Lemoine said LaMDA had been \textbf{telling me what I wanted to hear}. ''}
\end{quote}

Again, this implies a complex relationship behind the model. It implies that the model has understood what the correspondent expects from it and has deliberately conformed to that style. It implies that LaMDA has generated an internal model of the user's mind and has generated responses in accordance with its understanding of the agency of the user. A more rational explanation is that the model produces the most likely response, given a set of prior prompts.

It is clear that there is a significant difference in the level of anthropomorphisation used in scholarly literature and that used in journalistic reporting of that scholarly literature. However, an accurate reporting on AI is equally (if not more) important in case of popular media compared to scientific writing, and we must expect accurate understanding of model capabilities from both.

\section*{Harms Arising}

We have offered perspectives on LLM consciousness and demonstrated that claims of consciousness are part of a wider trend of anthropomorphisation in reporting of LLM capabilities. There is significant potential for harms arising from assuming a model possess consciousness and thereby misunderstanding its capabilities. Misunderstanding of LLMs comes in two catgeories: underestimation and overestimation. In this section we outline some of the potential harms arising from each.

Overestimation of the power of LLMs (often referred to as `LLM hype') is currently pervasive in the media, where numerous AI startups, influencers and members of the public make spurious claims as to the abilities of LLMs, either now or in the future. Unreliable claims of LLM capabilities are prone to attract investment and effort that could have been better prioritised. This is an issue for researchers and for businesses alike, where financial and time constraints limit capacity for investment. Selecting the right technology for a given task is vitally important for success, yet non-experts cannot be expected to understand the technology if researchers within the field do not appropriately report the technology's capacity.

Whilst LLMs are highly capable, they are also expensive to run and generate significant emissions. An LLM could be used to write a simple well-formed email, but the author could also rely on simple, computationally inexpensive, rule-based tools for spelling and grammar correction. An LLM could be used to generate a list of ideas or instructions for a well-known task, but these could also be easily found via an internet search. An LLM could be used to format a piece of code, but so could the inbuilt tools in an IDE at much lower cost. These are all clear overuses of a powerful technology which are easily solvable at a lower financial and carbon cost without it. To distinguish between appropriate and inappropriate use cases for an LLM, users need a proportionate understanding of the technology.

Overestimation of LLM capabilities is also harmful as it leads to over-reliance on the models. Take for example a recent case of a lawyer submitting LLM-generated evidence to court. The evidence contained multiple references to non-existent cases - the results of LLM hallucination\footnote{\url{https://www.nytimes.com/2023/06/08/nyregion/lawyer-chatgpt-sanctions.html}}. In mitigation, the lawyer claimed that he ``did not comprehend that ChatGPT could fabricate cases''. Fundamentally, he did not understand that the technology is capable of `infactuality', over-relying on the model to generate credible information. Whilst this case has gained prominence due to its wide reporting in the media, there are certainly countless other minor cases of over-reliance on LLM capabilities leading to fake information.  

This over-reliance can be attributed in part to the users of LLMs false perception that the model is in some way similar to a human, possessing consciousness, free will and thought.  Explicit anthropomorphic claims lead users to assume that the ability of an LLM to generate believable text also implies other human-level abilities. In reality, this is not the case and a non-anthropomorphised explanation of the models abilities can be given to mitigate harms arising from overestimation.

On the other hand, people who underestimate LLM capabilities are equally prone to harms arising from a misunderstanding of the technology. Those that dismiss the technology as another fleeting technology fashion trend are incapable of distinguishing LLM generated text, and hence prone to falsely believe fake information generated by the model. In a world of language models, we must learn to rigorously fact check information sources, where we cannot be certain of the provenance of the information. Even Wikipedia articles may now be contaminated with LLM text\footnote{https://slate.com/technology/2023/01/chatgpt-wikipedia-articles.html}.  

Bad actors will use large language models and applications powered by them to automate malicious behaviours. Adapting to this new reality is vital for information security at both the personal, and domestic levels. The advent of technology that can be used to automate communication either via automated email responses or via phone calls is potentially damaging for vulnerable members of society who may not realise that the agent they are communicating with is in fact an LLM.

Finally, underestimation of LLM capabilities may be harmful for those that do not adopt the technology into their working practices. Whilst others progress more quickly and efficiently due to using LLMs to automate their businesses, those that do not will be stuck using less efficient technologies. A right estimation and adoption of LLM technologies for appropriate tasks is key to harness their efficiency.


\section*{Study Limitations} \label{sec:limitations}



\subsection*{Model Consciousness}

Our study presupposes that it is valuable to distinguish between consciousness and non-consciousness in regard to LLMs such as LaMDA. A critic may argue that such discourse detracts from the progress of the field, or that the conclusions can only ever be subjective. A clear purpose of making this distinction comes in one of the key calls surrounding LLMs, to involve them as a participant in research studies. For example, Blake Lemoine claims that Google, in using LaMDA as property, violates the 13th amendment to the US constitution, abolishing slavery. He also maintains LaMDA `hired' a lawyer to represent it in the dispute\footnote{\url{https://www.wired.com/story/blake-lemoine-google-lamda-ai-bigotry/}}. AI-generated art, even though it is currently based on images rather than text, is a good example of a legal discussion around human-like capabilities of algorithms \citep{Epstein2020}. Similarly, there has been a recent trend towards attributing co-authorship of scholarly work to ChatGPT \citep{o66chatgpt,transformer2022rapamycin}, swiftly banned or discouraged by publishers\footnote{\url{https://2023.aclweb.org/blog/ACL-2023-policy/}, \url{https://icml.cc/Conferences/2023/llm-policy}}. We posit that by demonstrating the barriers that exist to the potential consciousness of LaMDA and other such models, we are able to speak into the ethical discussion of LLM use.

We have analysed an LLM (namely LaMDA) from the point of view of IIT, since among the theories of consciousness, it appears the most applicable to neural networks. Among other attempts of explaining consciousness, some rely on non-algorithmic processing \citep{Penrose1994}, e.g. implemented through quantum mechanics \citep{Hameroff1994}. But they even more clearly disqualify consciousness in LLMs, which are purely deterministic automata. However, other approaches are possible. For example, Ag\"uera y Arcas \citep{AgueraYArcas2022} analyses LaMDA through the lens of social and attentional theory of consciousness.

Further, we discussed whether LLMs can be conscious. However, many other important questions may also be asked about an AI's `mental capabilities', such as: Does it understand the language? Does its understanding have similarity with ours? Even more generally: is it intelligent?, or, Does it think? These questions are different and may be tackled with different approaches. For the most prominent example, see answering the latter question by observing machine's answers in a seminal historical piece on machine intelligence by Turing  \citep{turing1950}.

\subsection*{Generalisability}

While we used LaMDA as a focus point in our study, the research on generative LLMs is extremely homogeneous. All popular modern LLMs are based on the Transformer architecture \citep{vaswani2017attention}. This means that our conclusions, especially regarding consciousness, apply to other models such as BERT \citep{devlin-etal-2019-bert}, GPT-3 \citep{NEURIPS2020_1457c0d6}, T5 \citep{raffel2020exploring} and future models using the same, or similar, architectures.

For our conclusions to no longer apply, a model would need to overcome significant barriers to consciousness, such as continually processing and responding to signal information and acting as a cohesive unified architecture, rather than as a set of separately engineered units.

\subsection*{Subjectivity}


Our approach to identifying anthropomorphic language in both scholarly and journalistic reporting relies on the subjective application of the framework by one author of this work. Others may come to different conclusions in their application of the same framework to the same texts. These parts of the study are not aimed at precisely quantifying the levels of anthropomorphisation, but rather at highlighting its existence and giving a prompt for discussion on this topic within the AI community. The trend of overclaiming on LLM abilities is implicitly conditioned on the misuse of anthropomorphic language in scholarly and journalistic reporting.


Similarly, the interpretation of model responses is subjectively pareidolic and where one person may read abilities of causal inference, or tendencies towards intelligent thought, another may interpret these responses in a way that does not lead to these conclusions. We were aware of this in our analysis and have tried to offer an objective way of talking about model responses where possible, however the authors of this work acknowledge their inherent biases (i.e., non-belief in the model's consciousness) and cannot claim to have fully overcome these in this work.

\section*{Related Work}
\label{sec:relatedwork}
We are not the first authors to consider LLMs from the perspective of consciousness. In fact, Ag\"uera y Arcas \citep{AgueraYArcas2022} offers an insider view of the LaMDA model in an article that presents several transcripts of interactions, demonstrating its failures in cognitive ability as well as its successes, but offering no definite conclusion as to the possibility of consciousness in a LLM. More generally, other researchers have sought to apply cognitive frameworks to the problem of language modelling \citep{Pepperell2022,levin2022generalizing,chalmers2023large}, reaching similar conclusions to ours --- that a LLM cannot be considered conscious, based on the current evidence. An interesting perspective on the possibility of model consciousness is offered by Wei et al. \citep{wei2022emergent}, who posit that increasingly larger models appear to have emergent properties not possible in smaller models. One such emergent property in a future model may well be indistinguishable from consciousness.


For other discussions of the ability of machines to interpret language and the degree to which a machine may `understand' a text, we refer the reader to the recent work of Bender et al. \citep{bender-koller-2020-climbing,Bender2021} and Mitchell et al. \citep{mitchell2022debate}. In their work, an argument is built up towards the inability of a machine to attain human level recognition, with the important distinction between form and meaning being drawn. The popularised term of `stochastic parrots' was introduced to describe the functioning of LLMs as random sampling from the training data, with the human reader of the generated text playing a central role in assigning meaning.

A more sympathetic treatment of machine understanding is given in alternative literature, where researchers have claimed that LLMs are paving the way for General AI \citep{Manning2022}, or considered LLMs from the perspective of the human mind \citep{Lake2021}.


Regarding the anthropomorphic language describing AI, its usage has been recognised as potentially detrimental in other AI sub-fields, including machine learning \citep{Lipton2019}, brain-inspired AI \citep{Salles2020} or art-generating AI \citep{Epstein2020}.  Even for the AI techniques that were explicitly modelled to resemble human neurons, the similarity to actual human cognition has been described as overstated, possibly leading to dangerous misunderstandings \citep{Watson2019}. Looking more broadly, Brooker et al. \citep{Brooker2019} take a philosophical approach to highlight the error of assigning human-like qualities to a machine or an algorithm, but also thoroughly discusses its practical consequences. The main conclusion is that although using such language might seem a convenient shorthand, it may in fact hinder accurate understanding of the inner workings of the AI.

Similar to our analyses on anthropomorphic language in scientific reporting, 
Bender et al. \citep{bender-koller-2020-climbing} also report on the phenomenon of `Hype' in language modelling, analysing snippets from both scholarly and journalistic texts. They draw a similar conclusion to ours, that the ambiguous way researchers talk about LLMs is at risk of misleading the public about their capabilities. However, to the best of our knowledge, our study is the first to quantitatively assess the anthropomorphic language in AI/NLP scholarly literature. 



Finally, we note that there are a number of other important ethical concerns to be raised about the nature of language modelling beyond the dangers of false claims of consciousness. Importantly, work has focussed on the reproducibility of results in the LLM world, with a push towards the use of model cards to properly describe the underlying technology \citep{mitchell19model}. The training of large models requires immense computing resources, not only limiting the ability to investigate them to a few powerful research centers \citep{Luitse2021}, but also resulting in significant emissions outputs \citep{strubell-etal-2019-energy}, as does the conference system associated with reporting on these \citep{przybyla-shardlow-2022-using}. LLMs are also capable of producing falsehoods \citep{dziri-etal-2022-origin} or bias \citep{pmlr-v139-liang21a}. We must carefully address a range of important ethical issues \citep{Okerlund2022} alongside those raised in our work.

\section*{Conclusion} \label{sec:conclusion}

We have discussed the capacity of language models as conscious agents. We demonstrate that the capacities of modern transformer based LLMs, and specifically LaMDA, are not sufficient to constitute consciousness as defined by the framework introduced by Tononi et al. \citep{Tononi2015}. We further show that the forms of language used in modern NLP publications often refer to models in an anthropomorphic way. We argue that to address the problem, we must start at home and begin to use appropriate language.
Our investigations have focussed solely on models and examples in the English language, further studies into multilingual models, or monolingual models in languages other than English may yield additional insights beyond those we are able to give.
\color{black}
We demonstrate that the whilst researchers may unwittingly use ambiguous anthropomorphisms, this type of language is exacerbated in contemporary journalistic reports as explicit anthropomorphisation. We do not find sufficient evidence for the claims that the LaMDA model is conscious. In fact, we find that the claims of consciousness are the results of leading the model into responding as a human agent and incorrectly inferring anthropomorphic properties on the resulting generated sequences.
 
\noindent As a result of our findings, we make recommendations for the following three groups:
 
 \subsubsection*{Recommendations to Developers of LLMs}
 

Those that are developing and innovating with LLMs are a key group who must pay close attention to the style of language they use to describe their models.  They should consider using non-anthropomorphic language in the reporting of new innovations, and the capacity of models.

The following concrete actions could help achieve this objective:
\begin{description}
    \item[R1.1] Authors may include screening for anthropomorphisms as one of the stages of preparing their manuscript for publication. Among the phrases to look for are those used to describe mental capabilities of humans (e.g. \textit{intelligent}), especially using a model as a subject (e.g. \textit{model learns}).
    \item[R1.2] The authors should consider the audience that is expected to read a given document, and what they do and do not know about the topic \citep{Fischhoff2013}. It is especially important to avoid anthropomorphisms in highly influential contributions (e.g. LLMs intended for wide use) which are likely to be read by non-expert audiences.
    \item[R1.3] The contributors should also aim for maximum transparency when publishing their findings. The abilities of models, which are not publicly available, for example LaMDA \citep{thoppilan2022lamda}, DALL-E \citep{ramesh2021zero}, GPT-3 \citep{brown2020language}, ChatGPT and GPT-4 \citep{openai2023gpt4}, are virtually impossible to verify. 
    \item[R1.4] The above practices could be encouraged by journal editors and conference organisers. For example, the EACL organisers introduced a mandatory discussion of limitations\footnote{\url{https://2023.eacl.org/calls/papers/}} in 2023.
    \item[R1.5] We strongly advocate the use of human evaluation protocols for the outputs of LLMs. Whereas these models have been able to perform well on standard evaluation sets, they are known to produce troublesome text that cannot be picked up by automated evaluation, such as false or harmful information.
\end{description}

 
 
 
 

\subsubsection*{Recommendations to Users of LLMs}

Separate to those working on driving the innovation of ever larger and more capable LLMs, is the group of researchers and industry practitioners working with the models as consumers. The forefront of NLP research necessitates fine-tuning or prompting a large model. Those seeking to do so must be mindful of the model that they are choosing to use. The following might be helpful:

\begin{description}
    \item[R2.1] Gathering vital information about the model used, e.g. What source data was the model trained on? What is the modelling objective? How does the model handle vocabulary items? Does the model perform any filtering of its outputs?. Model cards \citep{mitchell19model} are a good starting point for this investigation.
    \item[R2.2] Considering how the choice of a model influences the ability to understand and replicate the research based on it. For example note that many models that are available via an API have never been published or released for scrutiny by the academic community.
\end{description}

\subsubsection*{Recommendations to Reporters of LLMs}


Finally, those reporting on progress in the LLM field in non-scholarly publications such as blogs, social media and the news must also be complicit in the effort to properly explain the capacities of LLMs. This is particularly difficult when the authors lack the technical expertise necessary for understanding a contribution.

Some techniques that could be used to overcome this include:
\begin{description}
    \item[R3.1] Being sceptical of the general statements made by authors in the introductory parts of the research papers. It is much better to rely on concrete examples or, whenever possible, first-hand interaction with a system.
    \item[R3.2] Partnering with AI researchers that can offer guidance -- preferably not the authors of the described contribution.
    \item[R3.3] Taking the status of the sources into account. Articles that have gone through peer review are less likely to include exaggerated claims than those appearing as preprints or those driven by commercial pressure.
\end{description}




\subsubsection*{Final Remarks}

We are a generation of researchers with the privilege to stand at a moment in time that will have a fundamental impact on the long-term future of our societies \citep{Greaves2021}. The recent growth in model capacity has led to machines capable of surpassing human performance on many tasks that were formerly thought impossible. We must choose to report on the capabilities of our models and machines responsibly and fairly. If reporting is left unchecked, the future of language modelling may be one where the capabilities of models are misunderstood. In overestimating the abilities, models are treated with more confidence than they deserve and given responsibilities they should not have. Misapplication leads to reinforcement of existing human biases and division in society. However, a future is also possible where responsible NLP/AI researchers have correctly communicated their findings and worked with journalists to explain the capabilities and limitations of these machines. Powerful models can be responsibly applied to great societal benefit.

Both these futures are possible, and the responsibility to shape the outcome lies with the AI community. Which way will we choose? 



%
%
%





\bibliography{tacl2021,anthology,PP,PP2,PP3,PP4}

\begin{thebibliography}{10}

\bibitem{devlin-etal-2019-bert}
Devlin J, Chang MW, Lee K, Toutanova K.
\newblock {BERT}: Pre-training of Deep Bidirectional Transformers for Language
  Understanding.
\newblock In: Proceedings of the 2019 Conference of the North {A}merican
  Chapter of the Association for Computational Linguistics: Human Language
  Technologies, Volume 1 (Long and Short Papers). Minneapolis, Minnesota:
  Association for Computational Linguistics; 2019. p. 4171--4186.
\newblock Available from: \url{https://aclanthology.org/N19-1423}.

\bibitem{NEURIPS2020_1457c0d6}
Brown T, Mann B, Ryder N, Subbiah M, Kaplan JD, Dhariwal P, et~al.
\newblock Language Models are Few-Shot Learners.
\newblock In: Larochelle H, Ranzato M, Hadsell R, Balcan MF, Lin H, editors.
  Advances in Neural Information Processing Systems. vol.~33. Curran
  Associates, Inc.; 2020. p. 1877--1901.
\newblock Available from:
  \url{https://proceedings.neurips.cc/paper/2020/file/1457c0d6bfcb4967418bfb8ac142f64a-Paper.pdf}.

\bibitem{brown2020language}
Brown T, Mann B, Ryder N, Subbiah M, Kaplan JD, Dhariwal P, et~al.
\newblock Language Models are Few-Shot Learners.
\newblock In: Larochelle H, Ranzato M, Hadsell R, Balcan MF, Lin H, editors.
  Advances in Neural Information Processing Systems. vol.~33. Curran
  Associates, Inc.; 2020. p. 1877--1901.
\newblock Available from:
  \url{https://proceedings.neurips.cc/paper/2020/file/1457c0d6bfcb4967418bfb8ac142f64a-Paper.pdf}.

\bibitem{raffel2020exploring}
Raffel C, Shazeer N, Roberts A, Lee K, Narang S, Matena M, et~al.
\newblock Exploring the limits of transfer learning with a unified text-to-text
  transformer.
\newblock J Mach Learn Res. 2020;21(140):1--67.

\bibitem{Bender2021}
Bender EM, Gebru T, McMillan-Major A, Schmitchell S.
\newblock {On the dangers of stochastic parrots: can language models be too
  big?}
\newblock In: Proceedings of FAccT '21. Association for Computing Machinery;
  2021. p. 271--278.

\bibitem{Okerlund2022}
Okerlund J, Klasky E, Middha A, Kim S, Rosenfeld H, Kleinman M, et~al.
\newblock {What's in the Chatterbox? Large Language Models, Why They Matter,
  and What We Should Do About Them}.
\newblock University of Michigan; 2022.
\newblock Available from:
  \url{https://stpp.fordschool.umich.edu/research/research-report/whats-in-the-chatterbox}.

\bibitem{Luitse2021}
Luitse D, Denkena W.
\newblock {The great transformer: Examining the role of large language models
  in the political economy of AI:}.
\newblock Big Data \& Society. 2021;8(2).
\newblock doi:{10.1177/20539517211047734}.

\bibitem{thoppilan2022lamda}
Thoppilan R, De~Freitas D, Hall J, Shazeer N, Kulshreshtha A, Cheng HT, et~al.
\newblock Lamda: Language models for dialog applications.
\newblock arXiv preprint arXiv:220108239. 2022;.

\bibitem{Tononi2015}
Tononi G, Koch C.
\newblock {Consciousness: here, there and everywhere?}
\newblock Philosophical Transactions of the Royal Society B: Biological
  Sciences. 2015;370(1668).
\newblock doi:{10.1098/RSTB.2014.0167}.

\bibitem{Jurafsky2021}
Jurafsky D, Martin JH.
\newblock {N-gram Language Models}.
\newblock In: Speech and Language Processing. Prentice Hall; 2021.

\bibitem{Austen1813}
Austen J.
\newblock {Pride and Prejudice}.
\newblock Whitehall, London: T. Egerton; 1813.

\bibitem{Markov2006}
Markov AA.
\newblock {An Example of Statistical Investigation of the Text Eugene Onegin
  Concerning the Connection of Samples in Chains}.
\newblock Science in Context. 2006;19(4):591--600.
\newblock doi:{10.1017/S0269889706001074}.

\bibitem{Smith2011}
Smith R.
\newblock {Limits on the application of frequency-based language models to
  OCR}.
\newblock In: Proceedings of the International Conference on Document Analysis
  and Recognition, ICDAR; 2011. p. 538--542.

\bibitem{Jelinek1976}
Jelinek F.
\newblock {Continuous Speech Recognition by Statistical Methods}.
\newblock Proceedings of the IEEE. 1976;64(4):532--556.
\newblock doi:{10.1109/PROC.1976.10159}.

\bibitem{Koehn2009}
Koehn P.
\newblock {Language Models}.
\newblock In: Statistical Machine Translation. Cambridge University Press;
  2009. p. 181--216.

\bibitem{Speaks2019}
Speaks J. {Theories of Meaning}; 2019.
\newblock Available from: \url{https://plato.stanford.edu/entries/meaning/}.

\bibitem{Pustejovsky2005}
Pustejovsky J. {Lexical Semantics: Overview}; 2005.

\bibitem{miller1990introduction}
Miller GA, Beckwith R, Fellbaum C, Gross D, Miller KJ.
\newblock Introduction to {W}ord{N}et: An on-line lexical database.
\newblock International journal of lexicography. 1990;3(4):235--244.

\bibitem{baker-etal-1998-berkeley}
Baker CF, Fillmore CJ, Lowe JB.
\newblock The {B}erkeley {F}rame{N}et Project.
\newblock In: {COLING} 1998 Volume 1: The 17th International Conference on
  Computational Linguistics; 1998.Available from:
  \url{https://aclanthology.org/C98-1013}.

\bibitem{Navigli2009a}
Navigli R.
\newblock {Word sense disambiguation}.
\newblock ACM Computing Surveys (CSUR). 2009;41(2):69.
\newblock doi:{10.1145/1459352.1459355}.

\bibitem{firth57}
Firth J.
\newblock A Synopsis of Linguistic Theory, 1930-55.
\newblock Oxford: Blackwell; 1957.

\bibitem{NIPS2013_9aa42b31}
Mikolov T, Sutskever I, Chen K, Corrado GS, Dean J.
\newblock Distributed Representations of Words and Phrases and their
  Compositionality.
\newblock In: Burges CJ, Bottou L, Welling M, Ghahramani Z, Weinberger KQ,
  editors. Advances in Neural Information Processing Systems. vol.~26. Curran
  Associates, Inc.; 2013.Available from:
  \url{https://proceedings.neurips.cc/paper/2013/file/9aa42b31882ec039965f3c4923ce901b-Paper.pdf}.

\bibitem{xing2014document}
Xing C, Wang D, Zhang X, Liu C.
\newblock Document classification with distributions of word vectors.
\newblock In: Signal and Information Processing Association Annual Summit and
  Conference (APSIPA), 2014 Asia-Pacific. IEEE; 2014. p. 1--5.

\bibitem{irsoy-cardie-2014-opinion}
{\. I}rsoy O, Cardie C.
\newblock Opinion Mining with Deep Recurrent Neural Networks.
\newblock In: Proceedings of the 2014 Conference on Empirical Methods in
  Natural Language Processing ({EMNLP}). Doha, Qatar: Association for
  Computational Linguistics; 2014. p. 720--728.
\newblock Available from: \url{https://aclanthology.org/D14-1080}.

\bibitem{passos-etal-2014-lexicon}
Passos A, Kumar V, McCallum A.
\newblock Lexicon Infused Phrase Embeddings for Named Entity Resolution.
\newblock In: Proceedings of the Eighteenth Conference on Computational Natural
  Language Learning. Ann Arbor, Michigan: Association for Computational
  Linguistics; 2014. p. 78--86.
\newblock Available from: \url{https://aclanthology.org/W14-1609}.

\bibitem{10.1145/2661829.2661974}
De~Vine L, Zuccon G, Koopman B, Sitbon L, Bruza P.
\newblock Medical Semantic Similarity with a Neural Language Model.
\newblock In: Proceedings of the 23rd ACM International Conference on
  Conference on Information and Knowledge Management. CIKM '14. New York, NY,
  USA: Association for Computing Machinery; 2014. p. 1819–1822.

\bibitem{vaswani2017attention}
Vaswani A, Shazeer N, Parmar N, Uszkoreit J, Jones L, Gomez AN, et~al.
\newblock Attention is All you Need.
\newblock In: Guyon I, Luxburg UV, Bengio S, Wallach H, Fergus R, Vishwanathan
  S, et~al., editors. Advances in Neural Information Processing Systems.
  vol.~30. Curran Associates, Inc.; 2017.Available from:
  \url{https://proceedings.neurips.cc/paper/2017/file/3f5ee243547dee91fbd053c1c4a845aa-Paper.pdf}.

\bibitem{Yang2019b}
Yang Z, Dai Z, Yang Y, Carbonell J, Salakhutdinov R, Le QV.
\newblock {XLNet: Generalized Autoregressive Pretraining for Language
  Understanding}.
\newblock arXiv: 190608237. 2019;.

\bibitem{Liu2019a}
Liu Y, Ott M, Goyal N, Du J, Joshi M, Chen D, et~al.
\newblock {RoBERTa: A Robustly Optimized BERT Pretraining Approach}.
\newblock arxiv. 2019;doi:{10.48550/arxiv.1907.11692}.

\bibitem{lewis-etal-2020-bart}
Lewis M, Liu Y, Goyal N, Ghazvininejad M, Mohamed A, Levy O, et~al.
\newblock {BART}: Denoising Sequence-to-Sequence Pre-training for Natural
  Language Generation, Translation, and Comprehension.
\newblock In: Proceedings of the 58th Annual Meeting of the Association for
  Computational Linguistics. Online: Association for Computational Linguistics;
  2020. p. 7871--7880.
\newblock Available from: \url{https://aclanthology.org/2020.acl-main.703}.

\bibitem{Clark2020a}
Clark K, Luong T, Le QV, Manning C.
\newblock ELECTRA: Pre-training Text Encoders as Discriminators Rather Than
  Generators.
\newblock In: Proceddings of ICLR; 2020.Available from:
  \url{https://openreview.net/pdf?id=r1xMH1BtvB}.

\bibitem{Church2022}
Church KW, Kordoni V.
\newblock {Emerging Trends: SOTA-Chasing}.
\newblock Natural Language Engineering. 2022;28(2):249--269.
\newblock doi:{10.1017/S1351324922000043}.

\bibitem{freitag-al-onaizan-2017-beam}
Freitag M, Al-Onaizan Y.
\newblock Beam Search Strategies for Neural Machine Translation.
\newblock In: Proceedings of the First Workshop on Neural Machine Translation.
  Vancouver: Association for Computational Linguistics; 2017. p. 56--60.
\newblock Available from: \url{https://aclanthology.org/W17-3207}.

\bibitem{jacobs-etal-2022-masked}
Jacobs CL, Hubbard RJ, Federmeier KD.
\newblock Masked language models directly encode linguistic uncertainty.
\newblock In: Proceedings of the Society for Computation in Linguistics 2022.
  online: Association for Computational Linguistics; 2022. p. 225--228.
\newblock Available from: \url{https://aclanthology.org/2022.scil-1.22}.

\bibitem{holtzmancurious}
Holtzman A, Buys J, Du L, Forbes M, Choi Y.
\newblock The Curious Case of Neural Text Degeneration.
\newblock In: International Conference on Learning Representations (ICLR);
  2020.

\bibitem{zhu2021retrieving}
Zhu F, Lei W, Wang C, Zheng J, Poria S, Chua TS.
\newblock Retrieving and reading: A comprehensive survey on open-domain
  question answering.
\newblock arXiv preprint arXiv:210100774. 2021;.

\bibitem{radford2018improving}
Radford A, Narasimhan K, Salimans T, Sutskever I, et~al.. Improving language
  understanding by generative pre-training; 2018.

\bibitem{radford2019language}
Radford A, Wu J, Child R, Luan D, Amodei D, Sutskever I. Language models are
  unsupervised multitask learners; 2019.

\bibitem{openai2023gpt4}
OpenAI. GPT-4 Technical Report; 2023.

\bibitem{chowdhery2022palm}
Chowdhery A, Narang S, Devlin J, Bosma M, Mishra G, Roberts A, et~al.
\newblock Palm: Scaling language modeling with pathways.
\newblock arXiv preprint arXiv:220402311. 2022;.

\bibitem{zhang2022opt}
Zhang S, Roller S, Goyal N, Artetxe M, Chen M, Chen S, et~al.
\newblock Opt: Open pre-trained transformer language models.
\newblock arXiv preprint arXiv:220501068. 2022;.

\bibitem{touvron2023llama}
Touvron H, Lavril T, Izacard G, Martinet X, Lachaux MA, Lacroix T, et~al..
  LLaMA: Open and Efficient Foundation Language Models; 2023.

\bibitem{ouyang2022training}
Ouyang L, Wu J, Jiang X, Almeida D, Wainwright C, Mishkin P, et~al.
\newblock Training language models to follow instructions with human feedback.
\newblock Advances in Neural Information Processing Systems.
  2022;35:27730--27744.

\bibitem{kwon2023reward}
Kwon M, Xie SM, Bullard K, Sadigh D. Reward Design with Language Models; 2023.

\bibitem{Tiku2022}
Tiku N. {Google engineer Blake Lemoine thinks its LaMDA AI has come to life};
  2022.
\newblock Available from:
  \url{https://www.washingtonpost.com/technology/2022/06/11/google-ai-lamda-blake-lemoine/}.

\bibitem{pereira}
{Pereira Jr } A.
\newblock {The Role of Sentience in the Theory of Consciousness and Medical
  Practice}.
\newblock Journal of Consciousness Studies. 2021;28(7-8):22--50.

\bibitem{Jones2013}
Jones RC.
\newblock {Science, sentience, and animal welfare}.
\newblock Biology \& Philosophy. 2013;28(1):1--30.
\newblock doi:{10.1007/s10539-012-9351-1}.

\bibitem{Nagel1974}
Nagel T.
\newblock {What Is It Like to Be a Bat?}
\newblock The Philosophical Review. 1974;83(4):435.
\newblock doi:{10.2307/2183914}.

\bibitem{Kirk2006}
Kirk R. {Zombies}; 2006.
\newblock \url{https://plato.stanford.edu/archives/sum2009/entries/zombies/}.

\bibitem{chalmers1995facing}
Chalmers DJ.
\newblock {Facing up to the problem of consciousness}.
\newblock Journal of consciousness studies. 1995;2(3):200--219.

\bibitem{Tononi2004}
Tononi G.
\newblock {An information integration theory of consciousness}.
\newblock BMC Neuroscience. 2004;5(1):1--22.
\newblock doi:{10.1186/1471-2202-5-42/COMMENTS}.

\bibitem{Mayner2018}
Mayner WGP, Marshall W, Albantakis L, Findlay G, Marchman R, Tononi G.
\newblock {PyPhi: A toolbox for integrated information theory}.
\newblock PLOS Computational Biology. 2018;14(7):e1006343.
\newblock doi:{10.1371/JOURNAL.PCBI.1006343}.

\bibitem{Kuzovkin2018}
Kuzovkin I, Vicente R, Petton M, Lachaux JP, Baciu M, Kahane P, et~al.
\newblock {Activations of deep convolutional neural networks are aligned with
  gamma band activity of human visual cortex}.
\newblock Communications Biology 2018. 2018;1(1):1--12.
\newblock doi:{10.1038/s42003-018-0110-y}.

\bibitem{Searle1980}
Searle JR.
\newblock {Minds, brains, and programs}.
\newblock Behavioral and Brain Sciences. 1980;3(3):417--424.
\newblock doi:{10.1017/S0140525X00005756}.

\bibitem{beltagy2020longformer}
Beltagy I, Peters ME, Cohan A.
\newblock Longformer: The long-document transformer.
\newblock arXiv preprint arXiv:200405150. 2020;.

\bibitem{anthropic2024claude}
Anthropic A. The claude 3 model family: Opus, sonnet, haiku; 2024.
\newblock Available from:
  \url{https://www-cdn.anthropic.com/de8ba9b01c9ab7cbabf5c33b80b7bbc618857627/Model_Card_Claude_3.pdf}.

\bibitem{KhandelwalLJZL20}
Khandelwal U, Levy O, Jurafsky D, Zettlemoyer L, Lewis M.
\newblock Generalization through Memorization: Nearest Neighbor Language
  Models.
\newblock In: 8th International Conference on Learning Representations, {ICLR}
  2020, Addis Ababa, Ethiopia, April 26-30, 2020. OpenReview.net;
  2020.Available from: \url{https://openreview.net/forum?id=HklBjCEKvH}.

\bibitem{fu2022hungry}
Fu DY, Dao T, Saab KK, Thomas AW, Rudra A, Re C.
\newblock Hungry Hungry Hippos: Towards Language Modeling with State Space
  Models.
\newblock In: The Eleventh International Conference on Learning
  Representations; 2022.

\bibitem{poli2023hyena}
Poli M, Massaroli S, Nguyen E, Fu DY, Dao T, Baccus S, et~al.
\newblock Hyena hierarchy: Towards larger convolutional language models.
\newblock In: International Conference on Machine Learning. PMLR; 2023. p.
  28043--28078.

\bibitem{peng2023rwkv}
Peng B, Alcaide E, Anthony Q, Albalak A, Arcadinho S, Biderman S, et~al.
\newblock RWKV: Reinventing RNNs for the Transformer Era.
\newblock In: Findings of the Association for Computational Linguistics: EMNLP
  2023; 2023. p. 14048--14077.

\bibitem{lewis2020retrieval}
Lewis P, Perez E, Piktus A, Petroni F, Karpukhin V, Goyal N, et~al.
\newblock Retrieval-augmented generation for knowledge-intensive nlp tasks.
\newblock Advances in Neural Information Processing Systems.
  2020;33:9459--9474.

\bibitem{borgeaud2022improving}
Borgeaud S, Mensch A, Hoffmann J, Cai T, Rutherford E, Millican K, et~al.
\newblock Improving language models by retrieving from trillions of tokens.
\newblock In: International conference on machine learning. PMLR; 2022. p.
  2206--2240.

\bibitem{Boly2013}
Boly M, Seth AK, Wilke M, Ingmundson P, Baars B, Laureys S, et~al.
\newblock {Consciousness in humans and non-human animals: recent advances and
  future directions}.
\newblock Frontiers in Psychology. 2013;4(OCT):1--20.
\newblock doi:{10.3389/FPSYG.2013.00625}.

\bibitem{Penrose1994}
Penrose R.
\newblock {Shadows of the Mind: A Search for the Missing Science of
  Consciousness}.
\newblock Oxford University Press; 1994.

\bibitem{10.3389/fnsys.2021.629436}
Braun HA.
\newblock Stochasticity Versus Determinacy in Neurobiology: From Ion Channels
  to the Question of the “Free Will”.
\newblock Frontiers in Systems Neuroscience. 2021;15.
\newblock doi:{10.3389/fnsys.2021.629436}.

\bibitem{kraikivski2021implications}
Kraikivski P.
\newblock Implications of Noise on Neural Correlates of Consciousness: A
  Computational Analysis of Stochastic Systems of Mutually Connected Processes.
\newblock Entropy. 2021;23(5):583.

\bibitem{Roose2023}
Roose K. {Why a Conversation with Bing's Chatbot Left Me Deeply Unsettled};
  2023.
\newblock Available from:
  \url{https://www.nytimes.com/2023/02/16/technology/bing-chatbot-microsoft-chatgpt.html}.

\bibitem{Epley2007}
Epley N, Waytz A, Cacioppo JT.
\newblock {On Seeing Human: A Three-Factor Theory of Anthropomorphism}.
\newblock Psychological Review. 2007;114(4):864--886.
\newblock doi:{10.1037/0033-295X.114.4.864}.

\bibitem{Parsons2023}
Parsons K. {ChatGPT has entered the classroom — and teachers are woefully
  unprepared}; 2023.
\newblock Available from:
  \url{https://www.thetimes.co.uk/article/chatgpt-has-entered-the-classroom-and-teachers-are-woefully-unprepared-czdsd35pv}.

\bibitem{Staton2023}
Staton B. {Education companies' shares fall sharply after warning over
  ChatGPT}; 2023.
\newblock Available from:
  \url{https://www.ft.com/content/0db12614-324c-483c-b31c-2255e8562910}.

\bibitem{Falati2023}
Falati S. {How ChatGPT Challenges Current Intellectual Property Laws}; 2023.

\bibitem{Oremus2023}
Oremus W. {He wrote a book on a rare subject. Then a ChatGPT replica appeared
  on Amazon.}; 2023.
\newblock Available from:
  \url{https://www.washingtonpost.com/technology/2023/05/05/ai-spam-websites-books-chatgpt/}.

\bibitem{Epstein2020}
Epstein Z, Levine S, Rand DG, Rahwan I.
\newblock {Who Gets Credit for AI-Generated Art?}
\newblock iScience. 2020;23(9):101515.
\newblock doi:{10.1016/J.ISCI.2020.101515}.

\bibitem{o66chatgpt}
O’Connor S, ChatGPT.
\newblock Open artificial intelligence platforms in nursing education: Tools
  for academic progress or abuse.
\newblock Nurse Education In Practice. 2023;66.

\bibitem{transformer2022rapamycin}
ChatGPT GPtT, Zhavoronkov A.
\newblock Rapamycin in the context of Pascal’s Wager: generative pre-trained
  transformer perspective.
\newblock Oncoscience. 2022;9:82.

\bibitem{Hameroff1994}
Hameroff SR.
\newblock Quantum coherence in microtubules: A neural basis for emergent
  consciousness?
\newblock Journal of consciousness studies. 1994;1(1):91--118.

\bibitem{AgueraYArcas2022}
Ag\"uera~y Arcas B.
\newblock {Do Large Language Models Understand Us?}
\newblock Daedalus. 2022;151(2):183--197.

\bibitem{turing1950}
Turing AM.
\newblock COMPUTING MACHINERY AND INTELLIGENCE.
\newblock Mind. 1950;LIX(236):433--460.
\newblock doi:{10.1093/mind/LIX.236.433}.

\bibitem{Pepperell2022}
Pepperell R.
\newblock {Does Machine Understanding Require Consciousness?}
\newblock Frontiers in Systems Neuroscience. 2022;16:52.
\newblock doi:{10.3389/FNSYS.2022.788486/BIBTEX}.

\bibitem{levin2022generalizing}
Levin M.
\newblock Generalizing frameworks for sentience beyond natural species.
\newblock Animal Sentience. 2022;7(32):15.

\bibitem{chalmers2023large}
Chalmers DJ. Could a Large Language Model be Conscious?; 2023.

\bibitem{wei2022emergent}
Wei J, Tay Y, Bommasani R, Raffel C, Zoph B, Borgeaud S, et~al.
\newblock Emergent Abilities of Large Language Models.
\newblock Transactions on Machine Learning Research. 2022;.

\bibitem{bender-koller-2020-climbing}
Bender EM, Koller A.
\newblock Climbing towards {NLU}: {On} Meaning, Form, and Understanding in the
  Age of Data.
\newblock In: Proceedings of the 58th Annual Meeting of the Association for
  Computational Linguistics. Online: Association for Computational Linguistics;
  2020. p. 5185--5198.
\newblock Available from: \url{https://aclanthology.org/2020.acl-main.463}.

\bibitem{mitchell2022debate}
Mitchell M, Krakauer DC.
\newblock The Debate Over Understanding in {AI}'s Large Language Models.
\newblock arXiv preprint arXiv:221013966. 2022;.

\bibitem{Manning2022}
Manning CD.
\newblock {Human Language Understanding \& Reasoning}.
\newblock Daedalus. 2022;151(2):127--138.

\bibitem{Lake2021}
Lake BM, Murphy GL.
\newblock {Word meaning in minds and machines.}
\newblock Psychological Review. 2021;doi:{10.1037/REV0000297}.

\bibitem{Lipton2019}
Lipton ZC, Steinhardt J.
\newblock {Troubling Trends in Machine Learning Scholarship}.
\newblock Queue. 2019;17(1).
\newblock doi:{10.1145/3317287.3328534}.

\bibitem{Salles2020}
Salles A, Evers K, Farisco M.
\newblock {Anthropomorphism in AI}.
\newblock AJOB Neuroscience. 2020;11(2):88--95.
\newblock doi:{10.1080/21507740.2020.1740350}.

\bibitem{Watson2019}
Watson D.
\newblock {The Rhetoric and Reality of Anthropomorphism in Artificial
  Intelligence}.
\newblock Minds and Machines. 2019;29(3):417--440.
\newblock doi:{10.1007/S11023-019-09506-6/FIGURES/6}.

\bibitem{Brooker2019}
Brooker P, Dutton W, Mair M.
\newblock {The new ghosts in the machine: 'Pragmatist' AI and the conceptual
  perils of anthropomorphic description}.
\newblock Ethnographic studies. 2019;16:272--298.
\newblock doi:{10.5281/ZENODO.3459327}.

\bibitem{mitchell19model}
Mitchell M, Wu S, Zaldivar A, Barnes P, Vasserman L, Hutchinson B, et~al.
\newblock Model Cards for Model Reporting.
\newblock In: Proceedings of the Conference on Fairness, Accountability, and
  Transparency. FAT* '19. New York, NY, USA: Association for Computing
  Machinery; 2019. p. 220–229.
\newblock Available from: \url{https://doi.org/10.1145/3287560.3287596}.

\bibitem{strubell-etal-2019-energy}
Strubell E, Ganesh A, McCallum A.
\newblock Energy and Policy Considerations for Deep Learning in {NLP}.
\newblock In: Proceedings of the 57th Annual Meeting of the Association for
  Computational Linguistics. Florence, Italy: Association for Computational
  Linguistics; 2019. p. 3645--3650.
\newblock Available from: \url{https://aclanthology.org/P19-1355}.

\bibitem{przybyla-shardlow-2022-using}
Przyby{\l}a P, Shardlow M.
\newblock Using {NLP} to quantify the environmental cost and diversity benefits
  of in-person {NLP} conferences.
\newblock In: Findings of the Association for Computational Linguistics: ACL
  2022. Dublin, Ireland: Association for Computational Linguistics; 2022. p.
  3853--3863.
\newblock Available from: \url{https://aclanthology.org/2022.findings-acl.304}.

\bibitem{dziri-etal-2022-origin}
Dziri N, Milton S, Yu M, Zaiane O, Reddy S.
\newblock On the Origin of Hallucinations in Conversational Models: Is it the
  Datasets or the Models?
\newblock In: Proceedings of the 2022 Conference of the North American Chapter
  of the Association for Computational Linguistics: Human Language
  Technologies. Seattle, United States: Association for Computational
  Linguistics; 2022. p. 5271--5285.
\newblock Available from: \url{https://aclanthology.org/2022.naacl-main.387}.

\bibitem{pmlr-v139-liang21a}
Liang PP, Wu C, Morency LP, Salakhutdinov R.
\newblock Towards Understanding and Mitigating Social Biases in Language
  Models.
\newblock In: Meila M, Zhang T, editors. Proceedings of the 38th International
  Conference on Machine Learning. vol. 139 of Proceedings of Machine Learning
  Research. PMLR; 2021. p. 6565--6576.
\newblock Available from:
  \url{https://proceedings.mlr.press/v139/liang21a.html}.

\bibitem{Fischhoff2013}
Fischhoff B.
\newblock {The sciences of science communication}.
\newblock Proceedings of the National Academy of Sciences.
  2013;110(supplement\_3):14033--14039.
\newblock doi:{10.1073/PNAS.1213273110}.

\bibitem{ramesh2021zero}
Ramesh A, Pavlov M, Goh G, Gray S, Voss C, Radford A, et~al.
\newblock Zero-shot text-to-image generation.
\newblock In: International Conference on Machine Learning. PMLR; 2021. p.
  8821--8831.

\bibitem{Greaves2021}
Greaves H, MacAskill W.
\newblock {The case for strong longtermism}.
\newblock Global Priorities Institute; 2021.
\newblock Available from:
  \url{https://globalprioritiesinstitute.org/hilary-greaves-william-macaskill-the-case-for-strong-longtermism-2/}.

\end{thebibliography}

\appendix
\section{ACL Abstracts Anthropomorphism Analysis}
\label{app:100abstracts}

In support of developing an understanding of anthropomorphism and quantifying the scale of the phenomena, we analysed 100 abstracts from the 60th Annual Meeting of the Association for Computational Linguistics (ACL 2022). For each abstract we identified whether the abstract mentioned LLMs (90/100) and whether the abstract contained any use of anthropomorphic language (AL) (36/100). Our findings in full are below. We have included the ACL anthology ID in each instance. 

\begin{table}[!ht]
    \tiny
    \centering
    \begin{tabular}{|l|c|c|p{7cm}|}
    \hline
        \textbf{ACL ID} & \textbf{LLM} & \textbf{AL} & \textbf{Examples} \\ \hline
        2022.acl-long.1 & y & n & ~ \\ \hline
        2022.acl-long.2 & n & n & ~ \\ \hline
        2022.acl-long.3 & y & y & AGG addresses the degeneration problem \\ \hline
        2022.acl-long.4 & y & n & ~ \\ \hline
        2022.acl-long.5 & y & y & incorporated into NDR models to perform the imagination of unseen counterfactual \\ \hline
        2022.acl-long.6 & y & y & it creates a way of smoothing \\ \hline
        2022.acl-long.7 & y & n & ~ \\ \hline
        2022.acl-long.8 & y & y & success of reinforcement learning (RL) in solving complex tasks, its capacity to explore and exploit an environment \\ \hline
        2022.acl-long.9 & y & n & ~ \\ \hline
        2022.acl-long.10 & y & n & ~ \\ \hline
        2022.acl-long.11 & y & y & easier to learn for student models, summaries produced by our students are shorter and more abstractive \\ \hline
        2022.acl-long.12 & y & n & ~ \\ \hline
        2022.acl-long.13 & y & n & ~ \\ \hline
        2022.acl-long.14 & y & y & cross-lingual distillation models merely conside \\ \hline
        2022.acl-long.15 & y & y & they still struggle to address \\ \hline
        2022.acl-long.16 & y & n & ~ \\ \hline
        2022.acl-long.17 & n & n & ~ \\ \hline
        2022.acl-long.18 & y & n & ~ \\ \hline
        2022.acl-long.19 & y & y & A sparse attention matrix estimation module, which predicts dominant elements of an attention matrix based,  FSAT successfully learned, FSAT remarkably outperforms, achieves \\ \hline
        2022.acl-long.20 & y & y & explainable recommendation models learn to discover, Though able to provide plausible explanations, can we immerse the models in a multimodal environment to gain proper awareness of real-world concepts and alleviate above shortcomings? the explainable recommendation model is encouraged to visualize what it refers to \\ \hline
        2022.acl-long.21 & n & n & ~ \\ \hline
        2022.acl-long.22 & y & y & model predicts winners/losers of bills and then utilizes them \\ \hline
        2022.acl-long.23 & y & y & increasing difficulties in understanding the dialogue history for both human and machine. \\ \hline
        2022.acl-long.24 & y & y & a refined empathy analysis is needed for comprehending fragile and nuanced human feelings,  SDMPED achieve the state-of-the-art performance. \\ \hline
        2022.acl-long.25& y & y & they generally employ, most of them focus on expressing empathy, responds skillfully \\ \hline
        2022.acl-long.26 & y & n & ~ \\ \hline
        2022.acl-long.27 & y & n & ~ \\ \hline
        2022.acl-long.28 & y & y & we find that standard BERT fine-tuning can partially learn the correct relationship, \\ \hline
        2022.acl-long.29 & y & y & infers an answer, \\ \hline
        2022.acl-long.30 & y & n & ~ \\ \hline
        2022.acl-long.31 & y & n & ~ \\ \hline
        2022.acl-long.32 & y & n & ~ \\ \hline
        2022.acl-long.33 & y & n & ~ \\ \hline
        2022.acl-long.34 & y & y & models fine-grained learning skills., FairytaleQA is capable of asking high-quality and more diverse questions. \\ \hline
        2022.acl-long.35 & y & n & ~ \\ \hline
        2022.acl-long.36 & y & n & ~ \\ \hline
        2022.acl-long.37 & y & n & ~ \\ \hline
        2022.acl-long.38 & n & n & ~ \\ \hline
        2022.acl-long.39 & y & y & LTA trains an adaptive classifier \\ \hline
        2022.acl-long.40 & y & y & could understand tables better \\ \hline
        2022.acl-long.41 & y & y & could understand tables better \\ \hline
        2022.acl-long.42 & y & n & ~ \\ \hline
        2022.acl-long.43 & y & y & the problem of catastrophic forgetting \\ \hline
        2022.acl-long.44 & y & y & model tends to memorize the meta-training tasks while ignoring support sets when adapting to new tasks \\ \hline
        2022.acl-long.45 & y & n & ~ \\ \hline
        2022.acl-long.46 & y & n & ~ \\ \hline
        2022.acl-long.47 & y & n & ~ \\ \hline
        2022.acl-long.48 & y & y & locate the key event information that determines the judgment, and (2) exploit the cross-task consistency constraints \\ \hline
        2022.acl-long.49 & y & n & ~ \\ \hline
        2022.acl-long.50 & y & n & ~ \\ \hline

        \end{tabular}
    \end{table}

\begin{table}[!ht]
    \tiny
    \centering
    \begin{tabular}{|l|c|c|p{7cm}|}
    \hline
        \textbf{ACL ID} & \textbf{LLM} & \textbf{AL} & \textbf{Examples} \\ \hline
        2022.acl-long.51 & y & y & Transformer based re-ranking models can achieve high search relevance \\ \hline
        2022.acl-long.52 & y & n & ~ \\ \hline
        2022.acl-long.53 & y & y & On BinaryClfs ICT improves the average AUC-ROC score by an absolute 10\% and reduces the variance due to example ordering by 6x and example choices by 2x. \\ \hline
        2022.acl-long.54 & y & n & ~ \\ \hline
        2022.acl-long.55 & y & y & Our proposed model named PRBoost achieves this goal via iterative prompt-based rule discovery and model boosting. It uses boosting to identify large-error instances and discovers candidate rules from them by prompting pre-trained LMs with rule templates. \\ \hline
        2022.acl-long.56 & y & n & ~ \\ \hline
        2022.acl-long.57 & n & n & ~ \\ \hline
        2022.acl-long.58 & y & n & ~ \\ \hline
        2022.acl-long.59 & y & n & ~ \\ \hline
        2022.acl-long.60 & y & n & ~ \\ \hline
        2022.acl-long.61 & y & n & ~ \\ \hline
        2022.acl-long.62 & y & n & ~ \\ \hline
        2022.acl-long.63 & n & n & ~ \\ \hline
        2022.acl-long.64 & y & y & The latter learns to detect task relations by projecting neural representations from NLP models to cognitive signals (i.e., fMRI voxels). \\ \hline
        2022.acl-long.65 & y & n & ~ \\ \hline
        2022.acl-long.66 & n & n & ~ \\ \hline
        2022.acl-long.67 & y & n & ~ \\ \hline
        2022.acl-long.68 & y & y & Typical generative dialogue models utilize the dialogue history to generate the response. \\ \hline
        2022.acl-long.69 & y & n & ~ \\ \hline
        2022.acl-long.70 & y & y & Multilingual pre-trained models are able to zero-shot transfer knowledge from rich-resource to low-resource languages \\ \hline
        2022.acl-long.71 & y & n & ~ \\ \hline
        2022.acl-long.72 & y & n & ~ \\ \hline
        2022.acl-long.73 & y & y & Specifically, the mechanism enables the model to continually strengthen its ability on any specific type by utilizing existing dialog corpora effectively. \\ \hline
        2022.acl-long.74 & y & y & Supervised parsing models have achieved impressive results on in-domain texts. \\ \hline
        2022.acl-long.75 & y & n & ~ \\ \hline
        2022.acl-long.76 & y & n & ~ \\ \hline
        2022.acl-long.77 & y & y & Transformers have been shown to be able to perform deductive reasoning on a logical rulebase containing rules and statements written in natural language. \\ \hline
        2022.acl-long.78 & y & n & ~ \\ \hline
        2022.acl-long.79 & y & n & ~ \\ \hline
        2022.acl-long.80 & y & y & A desirable dialog system should be able to continually learn new skills without forgetting old ones \\ \hline
        2022.acl-long.81 & y & n & ~ \\ \hline
        2022.acl-long.82 & y & y & FORTAP outperforms state-of-the-art methods \\ \hline
        2022.acl-long.83 & y & n & ~ \\ \hline
        2022.acl-long.84 & n & n & ~ \\ \hline
        s2022.acl-long.85 & y & n & ~ \\ \hline
        2022.acl-long.86 & y & y & SemAE uses dictionary learning to implicitly capture semantic information from the review text and learns a latent representation of each sentence over semantic units. \\ \hline
        2022.acl-long.87 & y & n & ~ \\ \hline
        2022.acl-long.88 & y & y & Empirical results show TBS models outperform end-to-end and knowledge-augmented RG baselines \\ \hline
        2022.acl-long.89& y & n & ~ \\ \hline
        2022.acl-long.90 & y & n & ~ \\ \hline
        2022.acl-long.91 & y & n & ~ \\ \hline
        2022.acl-long.92 & y & n & ~ \\ \hline
        2022.acl-long.93 & y & n & ~ \\ \hline
        2022.acl-long.94 & n & n & ~ \\ \hline
        2022.acl-long.95 & y & y & By conducting comprehensive experiments, we show that the synthetic questions selected by QVE can help achieve better target-domain QA performance \\ \hline
        2022.acl-long.96 & y & n & ~ \\ \hline
        2022.acl-long.97 & y & n & ~ \\ \hline
        2022.acl-long.98 & y & n & ~ \\ \hline
        2022.acl-long.99 & n & n & ~ \\ \hline
        2022.acl-long.100 & y & n & ~ \\ \hline
    \end{tabular}
\end{table}

\end{document}